\documentclass[12pt]{article}
\usepackage{psfrag,amsmath,amssymb,latexsym,epic, eepicemu, epsfig,
 float, enumerate}
\usepackage[normalem]{ulem}
\usepackage{amsfonts}
\usepackage{graphicx}
\usepackage{epsfig,psfrag,amsmath,amssymb,latexsym, mathtools, color}
\usepackage{amscd}
\usepackage{amsfonts}
\usepackage{graphics}
\usepackage{graphicx}
\usepackage{float}
\usepackage[T1]{fontenc}
\numberwithin{equation}{section}

\def\a{\alpha}
\def\b{\beta}

\definecolor{Red}{rgb}{1.00, 0.00, 0.00}

\definecolor{Green}{rgb}{0.2, 0.5, 0.2}

\definecolor{Blue}{rgb}{0.00, 0.00, 1.00}

\def\university#1{{\sl \begin{center} #1 \vspace{2pt} \end{center} } }
\def\inst#1{\vspace{1pt} \unskip$^{#1}$}

\begin{document}
\title{MAP segmentation in Bayesian hidden Markov models: a case study}

\author{Alexey Koloydenko\inst{2}, Kristi Kuljus\inst{1}, J\"{u}ri Lember\inst{1}}

\maketitle
\university{\inst{1}University of Tartu, Estonia; \inst{2}Royal Holloway, University of London, UK}

\begin{abstract}
We consider the problem of estimating the maximum posterior probability (MAP) state sequence for a finite state and finite emission alphabet hidden Markov model (HMM) in the Bayesian setup, where both emission and transition matrices have Dirichlet priors. We study a training set consisting of thousands of protein alignment pairs. The training data is used to set the prior hyperparameters for Bayesian MAP segmentation. Since the Viterbi algorithm is not applicable any more, there is no simple procedure to find the MAP path, and several iterative algorithms are considered and compared. The main goal of the paper is to test the Bayesian setup against the frequentist one, where the parameters of HMM are estimated using the training data.
\end{abstract}


\section{Introduction}
Let $\mathbb{A}=\{a_1,\ldots,a_L\}$ be a finite alphabet.  Suppose we have a training set consisting of pairs of sequences $(x(k),y(k))$, $k=1,\ldots,m$, where $x(k)$ is a finite sequence of elements in $\mathbb{A}$, called observation sequence. The corresponding  sequence $y(k)$ has the same length as $x(k)$, but the elements of $y(k)$ belong to the state set $\{1,\ldots,K\}$. The sequences $x(k)$ can be of different length as $k$ varies. We assume that every pair $(x(k),y(k))$ is an  output of a hidden Markov model (HMM), possibly with different parameters. Thus, the studied parameters are a transition matrix $\mathbb{P}:K\times K$, an emission matrix $\mathbb{Q}: K\times L$ and a vector of initial probabilities $p_0:K\times 1$. Suppose that besides the training set we  observe another observation sequence $x$. We assume that $x$ is also generated by an HMM with some unknown parameter $\theta=(\mathbb{P},\mathbb{Q},p_0)$, and our goal is to estimate the corresponding MAP or Viterbi path. The problem would be trivial if we knew the parameters of the HMM that has generated $x$, instead we have only a training set. Observe that although the setup resembles a classical pattern recognition problem, this is not the case, because the number of possible classes ($K^{\rm length\,\, of\,\, sequence}$) is huge (also the length of sequences varies largely in the training set). Thus, the traditional pattern recognition methods are not applicable and one has to model the data instead. We assume that HMM is suitable for modeling the data.\\\\
In  general, there are three approaches to the  above problem of estimating the Viterbi path. In the case of frequentist approach we assume that all training sequences $(x(k),y(k))$
  are generated from the same HMM having a parameter $\theta^*$. Then also $x$ is an observation sequence from the same HMM (i.e. $\theta=\theta^*$) and the solution to our problem is
straightforward: estimate the unknown parameter from the training data, let the
estimate be $\hat{\theta}$.  Then apply the Viterbi algorithm to find
$\arg\max_s p(s|x,\hat{\theta})$, where $s$ stands for all possible state paths. In the  Bayesian approach we assume that the unknown
  parameter $\theta$ is generated by a prior $\pi$. When we know
  $\pi$, the best we can do is to find $\arg\max_s p(s|x)$, where
  $p(s|x)=\int p(s|x,\theta)p(d\theta|x)$. In the \textit{no training data} approach we do not believe that the training data are related to the parameter
  $\theta$, or we believe that the sequence $x$ is long enough to estimate $\theta$ solely based on $x$.
  In this case we can ignore the training data and apply the standard EM parameter estimation
 algorithm to $x$, obtain the parameter estimate $\hat{\theta}_{EM}$, and then apply the Viterbi algorithm to find
$\arg\max_s p(s|x,\hat{\theta}_{EM})$. \\\\
 While the first and third approaches are fairly easy to implement, the Bayesian one has several complications. The main issues are:
 a) how to determine the prior based on training data? b) how to find a path that maximizes $p(s|x)$?
We apply commonly used Dirichlet priors (see \cite{METRON,tag,johnson,garca,johnson2,two}) and, in the spirit of moment estimation, tune the hyperparameters so that the variance and expectation of the emission and transition probabilities under the prior
 match the respective empirical quantities in the training data. Such a choice of priors is sometimes called {\it empirical priors}. The construction of empirical priors is the content of Section \ref{sec:emp}.  The second issue of maximizing $p(s|x)$ is a serious optimization problem, because as discussed in Section \ref{sec:model}, under the priors on the transition matrix, the underlying Markov chain loses the Markov property, and under the priors on the emission parameters the observations are not conditionally independent any more.  Hence, we are not dealing with an HMM and there is no Viterbi algorithm to find  $\arg\max_sp(s|x)$. There exist several iterative algorithms to address this  maximization problem. To our best knowledge the four most commonly used algorithms are so-called {\it segmentation EM} (sEM), {\it segmentation MM} (sMM) and {\it variational Bayes} (VB) approach and {\it Bayesian EM} (BEM) method (see also \cite{METRON} and the references therein). All the four methods are iterative and not guaranteed to converge to the global maximum.  The sEM method is just the EM method, where the underlying state  path is taken as the parameter of interest and model parameters are considered \sout{as} nuisance parameters; sMM is so-called Viterbi training (also known as {\it classification EM} \cite{clEM1,clEM2}) that is notoriously wrong when it comes to parameter estimation \cite{AVT3,AVT4}, but that performs surprisingly well for our purposes; VB is just an application of variational Bayes optimization in HMM setting \cite{VBtutorial,VBtutorial2,titterington}; and BEM is just an application of the parametric EM algorithm, where the estimate $\hat{\theta}$ is found and the Viterbi algorithm is then applied with $\hat{\theta}$. The difference between BEM and the frequentist approach is that BEM uses prior information in the EM algorithm. The four algorithms are explicitly stated in Section \ref{sec:algorithms}, for their justification and properties we refer to \cite{METRON}. It should be mentioned that all the algorithms are sensitive to the choice of initial sequences, thus initial sequences should be chosen carefully.
\\\\
The main goal of the article is to compare the three approaches (frequentist, Bayesian and training data free) on real data. We use the protein secondary structure dataset, where $\mathbb{A}$ consists of 20 amino acids and where the underlying states $\{1,\ldots,6\}$ denote different types of foldings. We consider 1000 pairs as a training set and 1000 pairs as a test set. On the training set the parameters (in the frequentist approach) or hyperparameters (in the Bayesian approach) are determined. In the third, training data free approach, the training set is obviously not used. After determining the (hyper)parameters, we find the MAP estimate for every single $x$ in the test set and measure its goodness. In the case of  frequentist approach, the MAP path is obtained  just as the output of the Viterbi algorithm. In the case of Bayesian approach, the MAP path is obtained using VB and sEM algorithms.  Unlike sMM, those two algorithms are applicable for any set of hyperparameters, hence the choice. After obtaining a MAP sequence $\hat{y}$, we need to measure its quality. This is not so straightforward -- although for any test sequence $x$ we have the true state sequence $y$, the very tempting pointwise comparison of  $\hat{y}$ to $y$ (Hamming distance) is most certainly not the right criterion, because the MAP path does not minimize the expected number of errors. Recall that our goal is to find the Viterbi path that maximizes $p(s|x,\theta)$, where $\theta$ is the unknown parameter that generates the test pair $(x,y$). Therefore, the correct criterion for measuring goodness of $\hat{y}$ is $p(\hat{y}|x,\theta)$ -- the bigger the probability, the better performance.
Unfortunately $\theta$ is unknown. But since $y$ is known, a natural estimate of $\theta$ would be the empirical transition and emission matrices based on the pair $(x,y)$, let this estimate be denoted by $\tilde{\theta}$. The problem with $\tilde{\theta}$ is its  sparseness. Since the sequences are typically a few hundred letters long, the matrices obtained by a single pair are too sparse, thus most paths $\hat{y}$ would be inadmissable, because $p(\hat{y}|x,\tilde{\theta})=0$. Therefore, we take also into consideration  the  empirical priors obtained using the training set and calculate the posterior mean $\bar{\theta}=\int \theta p(d\theta|x,y)$. We combine $\tilde{\theta}$ with $\bar{\theta}$   to obtain eight different parameters, and each of them is used to measure the goodness of the estimated MAP paths. The approaches are tested in Section \ref{sec:approaches}. The results show that the Bayesian approach slightly outperforms the frequentist one, and the training data free approach totally fails even for relatively long sequences.\\\\
Since a MAP path in the Bayesian setup is not straightforward to find, in Section \ref{sec:testalg} we present a preliminary set of experiments to compare the performance of the four algorithms for finding the Bayesian MAP path. These examples study the following questions: 1) which is the best algorithm; 2) how to choose initial sequences and how sensitive are the algorithms with respect to initial sequences; 3) how do the hyperparameters influence the structure of output paths? The answers to these questions are of  interest  on their own, but also necessary for interpreting the results of the main experiments. To test the algorithms, we took a pair $(x,y)$ and used only this pair ( \textit{case 1a,b} experiments) as well as the whole dataset (\textit{case 2a,b}) to specify the hyperparameters. In particular, we used the data to specify transition and emission probabilities, and then we used several concentration parameters to determine the hyperparameters. In such a way we end up with a large set of priors. Any prior from the set determines an objective function $p(\cdot|x)$. The  difference with the main experiments in Section \ref{sec:approaches} is that the optimality criterion is now clearly defined. Using the objective functions we generated a set of 6000 initial sequences and ran our four iterative algorithms (sEM, sMM, VB and BEM) with these 6000 initial sequences. Every algorithm produced 6000 outputs (not all are different) and out of all output sequences we chose for every algorithm the one that maximized the criterion.
The results show that mostly sEM and sMM perform best and they act very similarly, their  similarity is briefly explained in Section \ref{sec:algorithms}. Since sEM is the only algorithm that is guaranteed to increase the objective function, it is also clear why they both perform best. The study demonstrates how the hyperparameters influence the structure of output sequences, and how sensitive the algorithms can be with respect to initial sequences. \\\\
Finally let us remark that the  empirical transition and emission matrices, even when estimated from a large training set, are  still typically rather sparse. This means that our priors are sparse as well. The sparseness is often an issue when Dirichlet models  such as this one are used, see \cite{LDA,tag,johnson2}. Therefore, we took the sparsity under  consideration from the very beginning -- we specified the impossible emissions and transitions in advance using the whole corpus, and we put the priors on non-zero entries only. This slightly complicates the notation but in Dirichlet models the sparsity is an issue that simply cannot be ignored.

\section{Model description}\label{sec:model}
\paragraph{Transition matrix.}
Let $K$ denote the number of underlying states (in our case study, $K=6$) and suppose we aim to model the dynamics of state evolution, called the underlying process.
One of the most standard approaches is to model the underlying process as a homogeneous Markov chain with $K\times K$  transition matrix  $\mathbb{P}=(p_{ij})$.
In practice it can often happen that some transitions are impossible. Then the corresponding entry of the transition matrix is zero, thus the whole matrix can be rather sparse. In our article we assume that these impossible transitions are known in advance and we keep these transition probabilities zero throughout the whole modeling process.
Let $K_i$ denote the number of non-zero transitions in row $i$ and let $p_{i,j(i)}$ denote the $j$-th non-zero element in row $i$. Thus, if
we know that $p_{i1}=0$, but $p_{i2}>0$, then $1(i)=2$ and $p_{i,1(i)}=p_{i2}$. Therefore, for row $i$, we only model the non-zero transitions
$(p_{i,1(i)},\ldots, p_{i,K_i(i)})$. Let the indices of these non-zero elements in row $i$ be given by the set $J(i):=\{1(i),\ldots,K_i(i)\}$, $i=1,\ldots,K$.\\\\
In the Bayesian setup the transition matrix is not known and  assumed to be random with some known prior distribution. There are many ways to specify the prior distribution, but the most common approach  \cite{koski,HMMbook,RobertMarin,corander,BayesNonparam, tag} is to  assume that the rows of a transition matrix are independent and the non-zero entries of the $i$-th row follow the Dirichlet  distribution:
 $$(p_{i,1(i)},\ldots, p_{i,K_i(i)})\big| \alpha_{i} \sim  \text{Dir}(\alpha_{i,1(i)},\ldots,\alpha_{i,K_i(i)}),\quad \alpha_{i,j(i)}>0.$$
  Thus, the prior distribution for a transition matrix is given by its non-zero entries as follows:
$$\pi \big(\mathbb{P}\big)=\pi(p_{1,1(1)},\ldots,p_{1,K_1(1)})\ldots
\pi (p_{K,1(K)},\ldots,p_{K,K_K(K)})\propto \prod_{i=1}^K\prod_{j \in J(i)} p_{ij}^{\alpha_{ij}-1},$$
provided $(p_{i,1(i)},\ldots,p_{i,K_i(i)})\in \mathbb{S}_{K_i}$, where
$\mathbb{S}_{K_i}$ is a unit simplex. Thus, if $i\to j$ is an
impossible transition and $\mathbb{P}$ is such that $p_{ij}>0$, then
$\pi(\mathbb{P})=0$.
\\\\
Given a state path $s:=(s_1,\ldots,s_n)\in \{1,\ldots, K\}^n$, let $n_{ij}(s)$ denote the number of transitions $i\to j$ in $s$: $n_{ij}(s)=\sum_{t=1}^{n-1}I_{i,j}(s_t,s_{t+1})$. Let $n_i(s)=\sum_j n_{ij}(s)$.
When the path $s$ has impossible transition(s), then for
every possible transition matrix its probability is zero, thus the
posterior distribution $p(\mathbb{P}|s)$ is not defined. Therefore, in what
follows, we consider only {\it admissible paths}, i.e. the
paths that satisfy $\sum_j n_{ij}(s)=\sum_{j\in J(i)} n_{ij}(s)\sout{.}$ { for all $i=1, 2, \ldots, K$.} Hence,
given an admissible path $s$, the posterior $p(\mathbb{P}|s)$ is given by
%
\[
p(\mathbb{P}|s) = \prod_{i=1}^K p\big((p_{i,1(i)},\ldots,p_{i,K_i(i)})|s\big) \propto \prod_{i=1}^K\prod_{j\in J(i)}p_{ij}^{\alpha_{ij}+n_{ij}(s)-1}.\]
Thus, the posterior $p(\mathbb{P}|s)$ is
such that the rows are independent and the $i$-th row has the Dirichlet distribution:
\begin{equation}\label{pos1}
 (p_{i,1(i)},\ldots,p_{i,K_i(i)})\big| s,\alpha_i \sim
\text{Dir}(\alpha_{i,1(i)}+n_{i,1(i)}(s),\ldots,\alpha_{i,K_i(i)}+n_{i,K_i(i)}(s)).\end{equation}
Throughout the paper we assume that the initial distribution $p_0$ is known.
Thus, $p_0$ is a fixed probability distribution over the state
space with $p_0(1)+\cdots+ p_0(K)=1$.
\paragraph{Losing the Markov property.} Given a state sequence $s$ and a transition matrix $\mathbb{P}$, the probability of
$s$ is given by (with $0^0=1$)
$$p(s|\mathbb{P})=p_0(s_1)\prod_{i=1}^K\prod_{j=1}^K p_{ij}^{n_{ij}(s)}.$$
Observe that for any inadmissible sequence the probability above is zero for any matrix  $\mathbb{P}$ that
belongs to the support of $\pi$. Thus, the  probability of any path $s$ under our Dirichlet prior is zero when $s$ is inadmissible, and
for admissible $s$ it is (see \cite{METRON})
\begin{equation}\label{ps}
p(s)=\int p(s|\mathbb{P})\pi(d\mathbb{P}) =p_0(s_1)\prod_{i=1}^K
\left[{\Gamma(|\alpha_i|)\over \Gamma(|\alpha_i|+n_i(s)|)}\prod_{j\in J(i)}
{\Gamma(\alpha_{ij}+n_{ij}(s))\over
\Gamma(\alpha_{ij})}\right],\end{equation}
where $\Gamma(0)/\Gamma(0):=1$ and $|\alpha_i|:=\sum_{j \in J(i)} \alpha_{ij}$.
It is very important to realize that the process $Y_1,Y_2,\ldots$ with finite-dimensional distributions specified by (\ref{ps})
is not a Markov chain any more. Moreover, the process has a  longer memory than a Markov chain.
Thus, in the Bayesian setup the model is certainly not a hidden Markov model any more.
To see that the process has a longer memory, observe that when $\alpha_{11}>0$, then the probability of the constant path $s=(1,\ldots,1)$ of length $n$ equals \sout{to}
\begin{align*}
p(1,\ldots,1)&=p_0(1){\Gamma(|\alpha_1|)\over \Gamma(\alpha_{11})}{\Gamma(\alpha_{11}+n-1)\over \Gamma(|\alpha_1|+n-1)}
\geq p_0(1){\Gamma(|\alpha_1|)\over \Gamma(\alpha_{11})}{\Gamma(\alpha_{11}+n-1)\over \Gamma(\alpha_{11}+k+n-1)}\\
&=p_0(1){\Gamma(|\alpha_1|)\over \Gamma(\alpha_{11})}\cdot {1\over (\alpha_{11}+n-1)(\alpha_{11}+n)\cdots (\alpha_{11}+n-2+k)},\end{align*}
where $k=\lceil (|\alpha_1|-\alpha_{11})\rceil$ (recall $\alpha_{ij}>0$, so $k\geq 1$). We see that the probability decreases polynomially, while in the case of a  Markov chain the same probability equals  $p_0(1)p_{11}^{n-1}$ and decreases exponentially. This means that under the measure in (\ref{ps}) the constant blocks have much more weight, and we shall observe this also in our numerical examples. The difference between a Markov chain and a
Markov chain under a Dirichlet prior can be explained best in terms of urns. Suppose we have $K$ urns of different colors, the urn of color $i$ containing $\alpha_{ij}$, $j\in J(i)$,   balls of color $j$ (all together $|\alpha_i|$ balls). The Markov chain with transition probabilities $p_{ij}={\alpha_{ij}/|\alpha_i|}$ can be described as follows: the first color (and $Y_1$) is chosen according to the initial distribution. Then a ball is drawn with replacement from the chosen urn. The color of the chosen ball determines the next urn and so on. All balls are drawn with replacement, thus the number of balls in every urn remains constant. Under a Dirichlet distribution, the process is the same except that balls are drawn in Polya's fashion: together with the chosen ball another ball of the same color is added. Thus, the total number of balls increases by one at every step. This extra ball is the Bayes influence that destroys the Markov property and significantly increases the memory of the process. Now it is also clear that when for every $i=1,\ldots,K$, $\alpha_{ii}\geq \alpha_{ij}$
for $j\in J(i)$ (as it is in our case study), then the path with maximum probability is the constant one, and the relative probability of the constant path $s=(i,\ldots,i)$ increases as $|\alpha_i|$ decreases (see Proposition 4.1 in \cite{METRON} and the discussion  thereafter). To recapitulate: in comparison to the Markov chain our measure $p$ in (\ref{ps}) puts  significantly heavier weights on paths that
are constant or contain long constant blocks.
\paragraph{Emission distributions.} Let $\mathbb{A}=\{a_1,\ldots,a_L\}$ (in our case study $L=20$) be the
emission alphabet, that is the set of possible observation values. For an HMM, given that $Y_t=i$, the observation $x_t$ is generated according to the probability distribution
 $q_{i1},\ldots,q_{iL}$ independently of everything else. Thus, the emission probability matrix $\mathbb{Q}=(q_{il})$ will be of size $K\times L$.
Again, the emission matrix can be sparse and we assume the zero elements -- {\it impossible emissions} -- are known. Let $L_i$ denote the number of letters in the alphabet that can be emitted from state $i$, and let $j(i)$ again denote the $j$-th letter
with positive emission probability for state $i$. For every state $i$, we model only the non-zero emissions $q_{i,1(i)},\ldots q_{i,L_i(i)}$, let the indices of these elements be given by
$E(i)=\{1(i),\ldots,L_i(i)\}$. \\\\
In the Bayesian setup the emission matrix is assumed to be random with some prior distribution. Again, we assume this distribution to be such that
the rows of the emission matrix are independent. We also assume that our model prior is such that the  emission and transition parameters are independent.
For every state $i$, the non-zero emission probabilities are distributed according to a Dirichlet prior
${\text {Dir}}(\beta_{i,1(i)},\ldots,\beta_{i,L_i(i)})$, therefore
$$\pi(\mathbb{Q})=\pi(q_{1,1(1)},\ldots,q_{1,L_1(1)})\ldots \pi(q_{L,1(L)},\ldots,q_{L,L_L(L)})\propto
\prod_{i=1}^K\prod_{l \in E(i)} (q_{il})^{\beta_{il}-1},$$
provided $(q_{i,1(i)},\ldots,q_{i,L_i(i)})\in \mathbb{S}_{L_i}$. Given a state path $s$ and an observation sequence $x\in \mathbb{A}^n$, we define
$$m_{il}(s,x):=\sum_{t=1}^n I_{i}(s_t)I_{a_l}(x_t),\quad
m_{i}(s):=\sum_{l}m_{il}(s,x)=\sum_{t=1}^n I_{i}(s_t).$$
Thus, $m_{il}(s,x)$ is the number of pairs $(i,a_l)$ in $(s,x)$.
We call a sequence pair $(s,x)$ {\it admissible} when $s$ is an admissible
path and $(s,x)$ has no impossible emissions: $\sum_{l=1}^L
m_{il}(s,x)=\sum_{l \in E(i)}m_{il}(s,x)$  for all $i=1,2,\ldots, K$. Thus,  given an admissible pair
$(s,x)$, the posterior $p(\mathbb{Q}|s,x)$ factorises as
\begin{equation}\label{pos2}
p(\mathbb{Q}|s,x)=\prod_{i=1}^K p\big((q_{i,1(i)},\ldots,
q_{i,L_i(i)})|s,x\big),\end{equation} where
$$(q_{i,1(i)},\ldots,q_{i,L_i(i)})\big|s,x,\beta_i \sim \text
{Dir}(\beta_{i,1(i)}+m_{i,1(i)}(s,x),\ldots,\beta_{i,L_i(i)}+m_{i,L_i(i)}(s,x)).$$
Since the priors on $\mathbb{Q}$ and $\mathbb{P}$ are independent, so are the posteriors, thus
$$p(\mathbb{Q},\mathbb{P}|s,x)=p(\mathbb{Q}|s,x)p(\mathbb{P}|s).$$
\paragraph{Log-likelihood.} Given the matrices $\mathbb{P}$ and $\mathbb{Q}$, for any state
sequence $s$ and observations $x$ the joint probability of the pair
$(s,x)$ can be calculated as
\begin{equation*}
p(s,x|\mathbb{P},\mathbb{Q})=p(s|\mathbb{P})p(x|s,\mathbb{Q})=p_0(s_1)\prod_{i=1}^K\prod_{j=1}^K
(p_{ij})^{n_{ij}(s)}\cdot \prod_{i=1}^K \prod_{l=1}^L
(q_{il})^{m_{il}(s,x)},\end{equation*} which equals 0 if $(s,x)$ is inadmissible.
With our priors, the  probability of an admissible pair
$(s,x)$ can be obtained by integrating over $\mathbb{P}$ and $\mathbb{Q}$ separately:
$$ p(s,x)=\int p(s,x|\mathbb{P},\mathbb{Q})\pi(d \mathbb{P})\pi(d\mathbb{Q})=\int
p(s|\mathbb{P})\pi(d\mathbb{P})\cdot \int
p(x|s,\mathbb{Q})\pi(d\mathbb{Q})=p(s)p(x|s),$$ where $p(s)$ is
given in (\ref{ps}) and $p(x|s)$ can be calculated as
\begin{equation}\label{pxs}
p(x|s)=\prod_{i=1}^{K} {\Gamma(|\beta_i|)\over
\Gamma(|\beta_i|+m_i(s)|)}\prod_{j\in L(i)}{\Gamma(\beta_{ij}+m_{ij}(s,x))\over
\Gamma(\beta_{ij})}.\end{equation}
For an inadmissible pair $(s,x)$ we have $p(s,x)=0$. With these formulas, $\ln p(x,s)=\ln p(s)+\ln p(x|s)$ can be calculated.\\\\
Again, it is important to observe that in the Bayesian setup, for a given state sequence $s$ the observations are not independent any more. In terms of Polya urns we now have  $K$ different urns and $L$ colors. The $i$-th urn  contains $\beta_{il}$ balls of color $l$, thus all together $|\beta_i|$ initial balls. Given a path $s$, the observations are generated by drawing balls from the urns in Polya's fashion (i.e. with replacement and {an extra ball added): when $s_t=i$, then a ball from the $i$-th urn is taken. Observe that such a model has a tendency to attach a letter to a specific state (urn). That is, if a certain letter, say $a_1$, has been observed quite often in the observation sequence $x$, then a path $s$ that attaches all such observations to a specific state, say 2 (i.e. $m_{2{\color{blue},}1}(s)=m_2(s)$), makes
 conditional probability $p(x|s)$  relatively big.
\\\\
To recapitulate: the two dimensional stochastic process $(X_1,Y_1),(X_2,Y_2),\ldots$ having a finite-dimensional distribution for any $n$ given by $p(x,y)=p(y)p(x|y)$, where $p(y)$ is the measure in (\ref{ps}) and $p(x|y)$
 is defined in (\ref{pxs}), has nothing to do with an HMM any more -- the $Y$-process is not a Markov chain and the observations $X$ are not conditionally independent. Thus the name "Bayesian HMM" is in this sense very misleading.
\\\\
Finally, let us remark that since in our case study the hyperparameters $\alpha$ and $\beta$ are estimated from the training corpus, the vectors $\beta_i$ (as well as $\alpha_i$) are very different as $i$ varies. This prevents the label switching problem \cite{spezia2009reversible}, meaning that the two paths $s$ and $s'$ obtained by switching  the labels yield the same probabilities: $p(x|s)=p(x|s')$ and are, therefore, equivalent under some permutation of $\{1,2,\ldots, K\}$.

\section{Empirical priors}\label{sec:emp}
Our training set consists of pairs $\{(x(k),y(k)\}$, $k=1,\ldots,m$,
where for every $k$, $x(k)$ is a sequence of observations and
$y(k)$ is the actual sequence of underlying states, so-called {\it true
path}. We assume that every pair is generated by an HMM with \sout{a} parameter $\theta_k$ which consists of the transition and emission matrix. We also assume that $\theta_1,\ldots,\theta_m$ are i.i.d. from some prior $\pi$.
The goal of the Bayesian segmentation is to use information  from the training set and perform
Bayesian segmentation for a test sequence $x$. By that we mean finding a sequence $\hat{y}$ that maximizes
\begin{equation}\label{Bayes}
p(s|x)=\int p(s|x,\theta)p(d\theta|x)
\end{equation}
(or equivalently $p(s)p(x|s)$) over all possible state paths $s$. Here $p(s)$ is as in (\ref{ps}) and $p(x|s)$ as in (\ref{pxs}). These functions
depend on hyperparameters $\alpha$ and $\beta$, and we shall now discuss how to choose them  using the training set.
\\\\
The first step is to determine impossible transitions and emissions according to the training set.
Let the number of all $i\to j$ transitions and the number of all $i\to l$ emissions for the sequence pair $\{(x(k),y(k)\}$ and for the whole
training set be denoted as
\[ n_{ij}(k):=n_{ij}(y(k)), \,\,\, m_{il}(k):=m_{il}(y(k),x(k)), \quad n_{ij}:=\sum_{k} n_{ij}(k), \,\,\, m_{il}:=\sum_{k} m_{il}(k).\]
Let
\[  n_i:=\sum_j n_{ij},\quad m_i:=\sum_l m_{il}.\]
A transition $i\to j$ is impossible if $n_{ij}=0$ and
an emission $i\to l$ is impossible if $m_{il}=0$. Thus a transition
(emission) is possible, if it is at least once encountered in the
training set. For impossible transitions
(emissions) we set $\a_{ij}=0$ ($\beta_{il}=0$) and these parameters will remain zero, the rest of
the hyperparameters must be strictly positive. Recall that for every
$i=1,\ldots,K$, we have denoted the number of possible transitions (emissions) for state $i$ by $K_i$ ($L_i$). \\\\
In a strictly Bayesian setup one could start with non-informative priors for both the transition and emission parameters:
$$\alpha_{i,1(i)}=\ldots= \alpha_{i,K_i(i)}=\beta_{i,1(i)}=\ldots=\beta_{i,L_i(i)}=1,\quad i=1,\ldots,K.$$
Using the information from the training set we can find the posterior distribution of the parameters:
\begin{align}
\label{pr1}
(p_{i,1(i)},\ldots,p_{i,K_i(i)})|y(1),\ldots,y(m)&\sim {\text {Dir}}(1+n_{i,1(i)},\ldots,1+n_{i,K_i(i)})\\
\label{pr2}
(q_{i,1(i)},\ldots,q_{i,L_i(i)})|x(1),\ldots,x(m),y(1),\ldots y(m)&\sim {\text {Dir}}(1+m_{i,1(i)},\ldots,1+m_{i,L_i(i)}).\end{align}
Those posteriors could then be considered  candidate priors for a test sequence $x$.
With candidate priors (\ref{pr1}) and (\ref{pr2}), the prior expectation and variance of every possible transition and emission parameter is given by
$$E(p_{ij})={n_{i{j}}+1\over n_i+K_i}=:p^*_{ij},\quad E(q_{il})={m_{il}+1\over m_i+L_i}=:q^*_{il},$$ and
$$\text{Var}(p_{ij})={(n_{ij}+1)(n_i+K_i-n_{ij}-1)\over (n_i+K_i)^2(n_i+K_i+1)}, \quad
  \text{Var}(q_{il})={(m_{il}+1)(m_i+L_i-m_{il}-1)\over (m_i+L_i)^2(m_i+L_i+1)}.$$
Since $n_i$ and $m_i$ can be very big for a big training set, the variances of the parameters can be very small, meaning that
the prior of $p_{ij}$ would be very much concentrated over the point
estimator $p^*_{ij}$ and the prior of $q_{il}$ would be concentrated over
the point estimator $q^*_{il}$. This means that the Viterbi path in the
Bayesian setup would be the same as the Viterbi path calculated
with matrices $(p^*_{ij})$ and $(q^*_{il})$ (the impossible
transitions and emissions in these matrices are zeros).
\\\\
In order to make it possible to tune the variances and vary the influence of the parameters'
prior distributions, we consider the following generalization of the prior distributions in
(\ref{pr1}) and (\ref{pr2}):
\begin{align}\label{pr3}
(p_{i,1(i)},\ldots,p_{i,K_i(i)})\sim {\text {Dir}}(N_i p^*_{i,1(i)},\ldots,N_i p^*_{i,K_i(i)})\\
\label{pr4}
(q_{i,1(i)},\ldots,q_{i,L_i(i)})\sim {\text {Dir}}(M_i q^*_{i,1(i)},\ldots,M_i q^*_{i,L_i(i)}),
\end{align}
where  $N_i$ and $M_i$, $i=1,\ldots,K$, are nonnegative numbers called
{\it concentration} or {\it precision} parameters. Thus, for a possible
transition (emission) we have the following prior parameters: $\a_{ij}=N_ip^*_{ij}$
($\b_{il}=M_iq^*_{il}$). Then the prior expectations are still
$p^*_{ij}$ and $q^*_{il}$, but the prior variances (for possible emissions and transitions) are now
$$\text{Var}(p_{ij})={ p^*_{ij}(1-p^*_{ij})\over N_i+1},\quad
\text{Var}(q_{il})={ q^*_{ij}(1-q^*_{il})\over M_i+1}.$$
Observe that $N_i=n_i+K_i$ and $M_i=m_i+L_i$ gives us the distributions in
(\ref{pr1}) and (\ref{pr2}).
\\\\
The choice of $N_i$ and $M_i$ is not so easy to make.
One possibility is to choose the concentration
parameters $N_i$ and $M_i$ so that the variances of the prior distributions correspond to the empirical variances.
For every sequence pair $\{(x(k),y(k)\}$ we can obtain empirical estimates of $p_{ij}$ and $q_{il}$  as $n_{ij}(k)/n_i(k)$ and
$m_{il}(k)/m_i(k)$, respectively, $k=1,\ldots,m$. It could be natural to expect that the prior variances of
$p_{ij}$ and $q_{il}$ are close to the weighted empirical variances of these estimates:
\begin{equation}\label{variances}
\widehat{\text{Var}(p_{ij})}:=\sum_{k=1}^m
w^t_i(k)\Big({n_{ij}(k)\over n_i(k)}-\hat{p}_{ij}\Big)^2,\quad
\widehat{\text{Var}(q_{il})}:=\sum_{k=1}^m
w^e_i(k)\Big({m_{il}(k)\over m_i(k)}-\hat{q}_{il}\Big)^2,\end{equation}
where
\begin{equation}\label{point-est}
w^t_i(k):={n_i(k)\over n_i},\quad w^e_i(k):={m_i(k)\over m_i},\quad
\hat{p}_{ij}:={n_{ij}\over n_i},
\quad \hat{q}_{il}:={m_{il}\over m_i}.
\end{equation}
The concentration parameters are then chosen so that the prior variances would be more or less equal to the quantities obtained by
(\ref{variances}). To find an $N_i$ such that $\text{Var}(p_{ij})=\widehat{\text{Var}(p_{ij})}$ for every $j$ would in general not be possible, because $N_i$ is the same for
each element in row $i$. Therefore, we sum the variances up over $j$ and $l$ to get the following equations:
$$\sum_j\widehat{\text{Var}(p_{ij})}=\sum_j\text{Var}(p_{ij})={1-\sum_j \big( p^*_{ij} \big)^2\over N_i+1},\quad
\sum_l\widehat{\text{Var}(q_{il})}=\sum_l\text{Var}(q_{il})= {1-\sum_l \big( q^*_{il} \big)^2\over M_i+1}.$$
The solutions are
\begin{equation}\label{MN}
N_i={1-\sum_j \big( p^*_{ij} \big)^2 \over \sum_j \widehat{\text{Var}(p_{ij})}}-1,
\quad M_i= {1-\sum_l \big( q^*_{il}\big)^2\over \sum_l \widehat{\text{Var}(q_{il})}}-1.
\end{equation}
Thus, in our experiments we consider priors (\ref{pr3}) and (\ref{pr4}), where $N_i$ and $M_i$ are determined by (\ref{MN}), we refer to these
priors as {\it empirical priors}. If the variances $\widehat{\text{Var}(p_{ij})}$ are small for example, then $N_i$ is very big,
meaning that the prior is heavily concentrated over $p_{ij}^*$.
\section{Bayesian segmentation algorithms}\label{sec:algorithms} After specifying the priors, the objective function $p(s|x)$ we want to maximize is fully defined.
Maximizing $p(s|x)$ is not a trivial task, because as discussed in Section \ref{sec:model}, the model we are dealing with now is not an HMM any more. The Markov property is lost and thus the Viterbi algorithm cannot be applied any more. The problem of maximizing $p(s|x)$ was closely examined in \cite{METRON}, where several iterative algorithms for maximizing it were studied. The experiments in \cite{METRON} show a reasonably good performance of four iterative algorithms: {\it segmentation EM (sEM)}, {\it segmentation MM (sMM)}, {\it variational Bayes approach (VB)} and classical {\it Bayesian EM parameter estimation (BEM)}. In this section,
we present these four algorithms explicitly for our model (in \cite{METRON} the model was different), but for the justification and theory behind them we refer the reader to \cite{METRON}.
All four algorithms have the following input parameters: observation sequence $x$, initial distribution $p_0$, forbidden transitions and emissions and
hyperparameters $\alpha=(\alpha_{ij})$ and $\beta=(\beta_{ij})$. Two algorithms -- sMM and BEM -- might not be applicable if there is a
non-zero hyperparameter that is strictly smaller than one. For applying BEM and VB algorithms we need to define expected number of transitions and smoothing probabilities.
\paragraph{Expected  frequency of transitions and smoothing probabilities.} Given a transition matrix $\mathbb{P}$, emission matrix $\mathbb{Q}$, initial distribution $p_0$ and
an observation sequence $x$, define a $K\times K$-matrix $(\xi_{ij})$ of expected  frequencies of transitions  $i\to j$} as follows:
$$\xi_{ij}:=\sum_{t=1}^{n-1}P(Y_t=i,Y_{t+1}=j|p_0,\mathbb{P},\mathbb{Q},x).$$
The smoothing probabilities are defined as
$$\gamma_t(i):=P(Y_t=i|p_0,\mathbb{P},\mathbb{Q},x),\quad i=1,\ldots,K, \quad t=1,\ldots, n.$$
All these probabilities can be calculated by the standard forward-backward algorithms.
\paragraph{Segmentation EM (sEM).}
Given an admissible pair $(s,x)$, define  $K\times K$ matrix $\mathbb{U}(s)=(u_{ij}(s))$ and  $K\times L$ matrix
$\mathbb{H}(s,x)=(h_{il}(s,x))$ as follows:
\begin{align*}
u_{ij}(s)&:=\left\{
           \begin{array}{ll}
            \exp[\psi(\alpha_{ij}+n_{ij}(s))-\psi(|\alpha_i|+n_i(s))], & \hbox{if $i\to j$ is a possible transition;} \\
             0, & \hbox{if $i\to j$ is an impossible transition.} \\
           \end{array}
         \right.
\\
h_{il}(s,x)&:=\left\{
           \begin{array}{ll}
           \exp[\psi(\beta_{il}+m_{il}(s,x))-\psi(|\beta_i|+m_i(s))], & \hbox{if $i\to l$ is a possible emission;} \\
             0, & \hbox{if $i\to l$ is an impossible emission.} \\
          \end{array}
         \right.
\end{align*}
Here $\psi$ stands for the digamma function.
\newline \newline
\textbf{Initalization}: start with an admissible sequence $s^{(0)}.$

\noindent \textbf{Iteration}:
\begin{itemize}
\item Given the sequence $s^{(r)}$, find the matrices $\mathbb{U}^{(r)}=\big(u_{ij}(s^{(r)})\big)$ and
$\mathbb{H}^{(r)}=\big(h_{il}(x,s^{(r)})\big)$.
\item
Given the transition matrix $\mathbb{U}^{(r)}=(u_{ij})$
and the emission matrix $\mathbb{H}^{(r)}=(h_{il})$, apply the Viterbi
algorithm to find
\begin{equation}\label{villa}
s^{(r+1)}:=\arg\max_s \Big(\ln p_{0,s_1}+\sum_{i,j}n_{ij}(s)\ln
u_{ij} +\sum_{i,l}m_{il}(s,x)\ln h_{il}\Big).\end{equation}
\item If $s^{(r+1)}=s^{(r)}$, then stop iteration.
\end{itemize}
\textbf{Output}: state sequence $s^{(r+1)}$.
\\\\
The algorithm is based on the observation that the maximization in (\ref{villa}) can be performed via usual
Viterbi algorithm even when
  the matrices $\mathbb{U}$ and $\mathbb{H}$ are not proper
transition and emission matrices, because typically $\sum_{j}u_{ij}
<1$ and $\sum_{l}h_{il}<1$ (see the proof of Lemma 2.1 in \cite{METRON}). The segmentation EM algorithm increases the
objective function at every iteration step: $p(s^{(r+1)}|x)\geq
p(s^{(r)}|x)$.
\paragraph{Segmentation MM (sMM).} This algorithm is guaranteed to
be applicable when for all possible transitions and emissions
$\alpha_{ij}\geq 1$ and $\beta_{il}\geq 1$.
Given an admissible pair $(s,x)$, define
the  $K\times K$ posterior mode transition matrix $\hat{\mathbb{P}}(s)=\big(\hat{p}_{ij}(s)\big)$ and the  $K\times L$ posterior
mode emission matrix $\hat{\mathbb{Q}}(s,x)=\big(\hat{q}_{il}(s,x)\big)$ as follows:
\begin{align*}
\hat{p}_{ij}(s)&:=\left\{
           \begin{array}{ll}
             {\a_{ij}+n_{ij}(s)-1\over \mid \a_i\mid+n_i(s)-K}, & \hbox{if $i\to j$ is a possible transition;} \\
             0, & \hbox{if $i\to j$ is an impossible transition.} \\
           \end{array}
         \right.
\\
\hat{q}_{il}(s,x)&:=\left\{
           \begin{array}{ll}
             {\b_{il}+m_{il}(s,x)-1\over \mid \b_i\mid+m_i(s,x)-K}, & \hbox{if $i\to l$ is a possible emission;}   \\
             0, & \hbox{if $i\to l$ is an impossible emission.} \\
           \end{array}
         \right.
\end{align*}
%
\textbf{Initalization}: start with an admissible sequence $s^{(0)}.$

\noindent \textbf{Iteration}:
\begin{itemize}
\item Given the sequence $s^{(r)}$, find the posterior mode transition matrix
$\mathbb{P}^{(r)}=\big(\hat{p}_{ij}(s^{(r)})\big)$ and posterior
mode emission matrix
$\mathbb{Q}^{(r)}=\big(\hat{q}_{il}(s^{(r)},x)\big)$.
\item Given the posterior mode matrices $\mathbb{P}^{(r)}=(\hat{p}_{ij})$
and $\mathbb{Q}^{(r)}=(\hat{q}_{il})$, apply the Viterbi algorithm to find
\begin{equation}\label{villa1}
s^{(r+1)}:=\arg\max_s \Big(\ln p_{0,s_1}+\sum_{i,j}n_{ij}(s)\ln
\hat{p}_{ij} +\sum_{i,l}m_{il}(s,x)\hat{q}_{il}\Big).
\end{equation}
\item If $s^{(r+1)}=s^{(r)}$, then stop iteration.
\end{itemize}
\textbf{Output}: sequence $s^{(r+1)}$.
\newline \newline
Let $\theta^{(r)}=(\mathbb{P}^{(r)},\mathbb{Q}^{(r)})$. This
algorithm increases the joint posterior likelihood:
$$p(\theta^{(r+1)},s^{(r+1)}|x)\geq p(\theta^{(r)},s^{(r+1)}|x)\geq
p(\theta^{(r)},s^{(r)}|x),$$ but not necessarily the marginal
posterior probability as sEM does. However, since for even moderate
$n$, $\psi(n)\approx \ln (n)$, we see that for big $n$,
$\hat{p}_{ij}\approx u_{ij}$ and $\hat{q}_{il}\approx h_{il}$, and
therefore, as we also see from our experiments,  sEM and sMM behave similarly.
\paragraph{Bayesian EM (BEM).} As in the case of sMM, this algorithm  is applicable when for all possible
transitions and emissions $\alpha_{ij}\geq 1$ and $\beta_{il}\geq 1$.
Given an observation sequence $x$, $(\xi_{ij})$ and $(\gamma_t(i))$, define the $K\times K$ transition and the  $K\times L$ emission matrix $\tilde{\mathbb{P}}=\big(\tilde{p}_{ij}\big)$ and
$\tilde{\mathbb{Q}}=\big(\tilde{q}_{il}\big)$  respectively as follows:
\begin{align*}
\tilde{p}_{ij}&:=\left\{
           \begin{array}{ll}
            (\a_{ij}+\xi_{ij}-1)/(|\a_i|+\sum_j \xi_{ij}-K), & \hbox{if $i\to j$ is a possible transition;} \\
             0, & \hbox{if $i\to j$ is an impossible transition.} \\
           \end{array}
         \right.
\\
\tilde{q}_{il}&:=\left\{
           \begin{array}{ll}
             (\b_{il}+ \sum_{t}\gamma_t(i)I_{a_l}(x_t)-1)/(\sum_{t}\gamma_t(i)+|\beta_i|-L) , & \hbox{if $i\to l$ is a possible emission;} \\
             0, & \hbox{if $i\to l$ is an impossible emission.} \\
           \end{array}
         \right.
\end{align*}
Observe that as in the case of the  posterior mode matrices, the sufficient conditions for parametric updates is that $\alpha_{ij}\geq
1$ for every possible transition and $\b_{il}\geq 1$ for every possible emission.
\newline \newline
\textbf{Initalization}: start with an admissible sequence $s^{(0)}$ and take for every $i,j$ and $t$,
$$\xi_{ij}^{(0)}:=n_{ij}(s^{(0)}),\quad \gamma^{(0)}_t(i)=I_i(s^{(0)}_t).$$
\textbf{Iteration}:
\begin{itemize}
\item Given  $(\xi_{ij}^{(r)})$ and $(\gamma_{t}^{(r)}(i))$, calculate the transition and emission matrices
$\mathbb{P}^{(r+1)}=\big(\tilde{p}_{ij}\big)$ and
$\mathbb{Q}^{(r+1)}=\big(\tilde{q}_{il}\big)$ by parametric update
formulas.
\item Apply the Viterbi algorithm with parameters
 $\mathbb{P}^{(r+1)}$ and $\mathbb{Q}^{(r+1)}$  to compute  the Viterbi path
\begin{equation*}
s^{(r+1)}:=\arg\max_s \Big(\ln p_{0,s_1}+\sum_{i,j}n_{ij}(s)\ln
\tilde{p}_{ij} +\sum_{i,l}m_{il}(s,x)\tilde{q}_{il}\Big).\end{equation*}
\item If $s^{(r+1)}=s^{(r)}$, stop iteration.
\item Apply the forward-backward recursions with $\mathbb{P}^{(r+1)}$ and $\mathbb{Q}^{(r+1)}$ to find the expected number of
transitions $\big(\xi_{ij}^{(r+1)}\big)$ and smoothing probabilities $\big(\gamma_{t}(i)^{(r+1)}\big)$ for the next iteration.
\end{itemize}
\textbf{Output}: sequence $s^{(r+1)}$.
\newline \newline
Bayesian EM is the usual parametric EM estimation algorithm in
the Bayesian setup. At every iteration step it increases the
posterior likelihood: let
$\theta^{(r)}=(\mathbb{P}^{(r)},\mathbb{Q}^{(r)})$, then
$p(\theta^{(r+1)}|x)\geq p(\theta^{(r)}|x)$ or equivalently,
$p(x|\theta^{(r+1)})\pi(\theta^{(r+1)})\geq
p(x|\theta^{(r)})\pi(\theta^{(r)})$. Observe also that calculating
$s^{(r+1)}$ at every iteration step is not actually necessary for
the algorithm  -- one could just  run the parametric EM algorithm
to obtain the final parameter estimate $\hat{\theta}$ and then
run the Viterbi algorithm to get the estimate $\hat{y}$. This would be
the traditional " first estimate the parameters, then perform segmentation" approach.
Since our goal is segmentation and not parameter estimation, we
calculate the Viterbi path at every iteration step and stop when the two consecutive state paths are equal. In practice, this approach typically reduces
the number of iterations, but there is a theoretical possibility for an
infinite loop. To avoid that, a maximum number of iterations should be specified.
\paragraph{VB algorithm.} Given an observation sequence $x$, $(\xi_{ij})$ and $(\gamma_t(i))$, define  $K\times K$ matrix
$\mathbb{U}$ and  $K\times L$ matrix $\mathbb{H}$ as follows:
\begin{align*}
u_{ij}&:=\left\{
           \begin{array}{ll}
            \exp[\psi(\alpha_{ij}+\xi_{ij})-\psi(|\alpha_i|+\sum_j \xi_{ij})], & \hbox{if $i\to j$ is a possible transition;} \\
             0, & \hbox{if $i\to j$ is an impossible transition.} \\
           \end{array}
         \right.
\\
h_{il}&:=\left\{
           \begin{array}{ll}
               \exp[\psi(\beta_{il}+\sum_{t}\gamma_t(i)I_{a_l}(x_t))-\psi(|\beta_i|+\sum_{t}\gamma_t(i))], &  \hbox{if $i\to l$ is a possible emission;} \\
             0, & \hbox{if $i\to l$ is an impossible emission.} \\
          \end{array}
         \right.
\end{align*}
\textbf{Initalization}: start with an admissible state sequence $s^{(0)}$ and take for every $i,j$ and $t$,
$$\xi_{ij}^{(0)}:=n_{ij}(s^{(0)}),\quad
\gamma^{(0)}_t(i)=I_i(s^{(0)}_t).$$
\textbf{Iteration}:
\begin{itemize}
\item Given $(\xi_{ij}^{(r)})$ and $(\gamma_{t}^{(r)}(i))$, calculate the parameter matrices $\mathbb{U}^{(r)}$
and $\mathbb{H}^{(r)}$.
\item Apply the Viterbi algorithm with $\mathbb{U}^{(r)}=(u_{ij})$ and $\mathbb{H}^{(r)}=(h_{il})$ to find
\begin{equation*}
s^{(r+1)}:=\arg\max_s \Big(\ln p_{0,s_1}+\sum_{i,j}n_{ij}(s)\ln
u_{ij} +\sum_{i,l}m_{il}(s,x)\ln h_{il}\Big).\end{equation*}
\item With $\mathbb{U}^{(r)}$
and $\mathbb{H}^{(r)}$ update $(\xi_{ij}^{(r+1)})$ and
$(\gamma_{t}^{(r+1)}(i))$, observe that even $\mathbb{U}^{(r)}$ and
$\mathbb{H}^{(r)}$ are not proper stochastic matrices.

\item If $s^{(r+1)}=s^{(r)}$, stop iteration.
\end{itemize}
\textbf{Output}: sequence $s^{(r+1)}$.
\\\\
As in the BEM case, calculating the sequence $s^{(r+1)}$ is needed for the
stopping rule, only. Also the number of iterations should be specified.

\section{Numerical experiments}
\subsection{Example 1: testing the segmentation algorithms}\label{sec:testalg}
\subsubsection{The first set of experiments} In our first set of experiments, we illustrate
the  behaviour of the four segmentation algorithms for a single
sequence pair $(x,y)$. Similar comparisons in the case of continuous
emission distributions were performed in \cite{METRON}. The
situation differs in comparison to the case studied in
\cite{METRON} regarding mainly two aspects: 1) we now have the discrete
case with Dirichlet emissions, 2) we have real data pairs $(x,y)$. The
general idea for testing the algorithms is the same as in
\cite{METRON}: we consider one sequence pair $(x,y)$, a set of
hyperparameters $\alpha$ and $\beta$, and a fixed set of initial
state sequences. After that we run all the four algorithms with all
initial sequences from our set and calculate $p(\hat{y}|x)$,
where $\hat{y}$ denotes the estimated Viterbi path for a given
initial sequence and given algorithm. Since the goal is to maximize
$p(y|x)$, the final Viterbi path estimate $\hat{v}$ for a considered
method and given hyperparameters is taken as  $\hat{v}=\arg
\max_{\hat{y}}p(\hat{y}|x)$, where maximum is taken over all
different output state sequences obtained for different initial sequences.
The method with biggest $p(\hat{v}|x)$ gives the best solution. In
\cite{METRON} it was observed that segmentation EM and segmentation
MM were often best.
\\\\
Observe that our goal is to maximize $p(\cdot|x)$ over all possible state
paths $y$. The function $p(\cdot|x)$ depends actually not only on
$x$, but also on the hyperparameters $\alpha$ and $\beta$, thus it
should be written as $p(\cdot|x,\alpha,\beta)$. As it is argued in
\cite{METRON} (see Section 3.5), the choice
of hyperparameters has a much larger impact on segmentation output
than the observation sequence $x$ itself, meaning that changing the
hyperparameters might drastically change the shape of the objective
function as well as the solution. Therefore, the second aim with the first experiments
is to demonstrate dependence of Bayesian segmentation
results on hyperparameters. In \cite{METRON}, the performance of the
algorithms was studied in simulation experiments, thus the
data-generating model was known. In the present case we do not know
the true model, but we know the true underlying state sequence $y$. In all examples, we consider a fixed pair $(x,y)$ from our data set with sequence length $n=327$, and a fixed initial distribution $p_0$ calculated from the whole training corpus. The hyperparameters $\alpha$ and $\beta$ will be factorized as $\alpha=N\mathbb{P}$ and
$\beta=M\mathbb{Q}$, where $M$ and $N$ are nonnegative concentration parameters, and $\mathbb{P}$ and $\mathbb{Q}$ are fixed transition and emission matrices. We shall consider four different sets of
matrices $\mathbb{P}$ and $\mathbb{Q}$, and for every set several different concentration parameters $M$ and $N$ will be considered. Recall that $\mathbb{P}$ and $\mathbb{Q}$ are prior expectations and decreasing concentration parameters increases the prior variances of the parameters.
\\\\
We will consider the following four cases in our first set of experiments:
\begin{description}
  \item[Case 1a] In the first analysis we find the estimates $\hat{\mathbb{P}}$ and $\hat{\mathbb{Q}}$ using the respective counts from only $(x,y)$. Observe that zero transitions and emissions are in this case determined by the pair $(x,y)$ and because of this there are many zeros in both matrices, especially in $\hat{\mathbb{Q}}$. The prior belief into $\hat{\mathbb{P}}$ and $\hat{\mathbb{Q}}$ can be tuned with different concentration parameters $N$ and $M$. Large constants $N$ and $M$ indicate a strong belief in our prior distributions, whereas small $N$ and $M$ give a smaller influence to prior distributions and larger influence to data. Fixing large $N$ or $M$ corresponds to fixing transition or emission parameters, respectively. Recall that when $N$ and $M$ are too small, then sMM and BEM could not be applicable. In order for them to be applicable, we need to guarantee that all non-zero transition and emission hyperparameters are larger than 1, thus we have to take $N>N_1:=(\min_{ij} \hat{p}_{ij})^{-1}$, $M>M_1:=(\min_{ij} \hat{q}_{ij})^{-1}$ (minimum is taken over non-zero entries).
  \item[Case 1b] In the second analysis we consider the case where the probability mass in each row of the parameter matrices $\hat{\mathbb{P}}$ and $\hat{\mathbb{Q}}$ is divided evenly between the non-zero entries. The zero entries are still specified by the same pair $(x,y)$ and they coincide with these of case 1a. Denote the respective parameter matrices by $\hat{\mathbb{P}}_0$ and $\hat{\mathbb{Q}}_0$, thus $\alpha=N \hat{\mathbb{P}}_0$, $\beta=M \hat{\mathbb{Q}}_0$.
  \item[Case 2a] In the third analysis we use information from 1000 training sequence pairs $(x,y)$ to calculate $\hat{\mathbb{P}}$ and $\hat{\mathbb{Q}}$. In comparison with case 1a and case 1b, the matrices are less sparse.
  \item[Case 2b] The emission and transition matrices $\hat{\mathbb{P}}_0$ and $\hat{\mathbb{Q}}_0$ have the same zeros as in case 2a, but the non-zero entries have the same value in every row, that is the probability mass in each row is uniformly distributed between the nonzero elements.
\end{description}
\paragraph{Initial sequences in the first experiments.} Since the output of the studied algorithms can be very sensitive with respect to initial state sequences, the choice of initial sequences is an important issue. Hence we consider many initial sequences, find the corresponding output sequences and choose the best of them as described above. Ideally a suitable set of initial sequences should somehow cover the whole search space. On the other hand, all sequences must be admissible. Since we have many zeros in the matrices $\hat{\mathbb{P}}$ and $\hat{\mathbb{Q}}$ (equally many zeros in $\hat{\mathbb{P}}_0$ and $\hat{\mathbb{Q}}_0$), the only plausible way to obtain admissible paths is to generate them from $p(\cdot|x,p_0,\hat{\mathbb{P}}, \hat{\mathbb{Q}})$ or from
$p(\cdot|x,p_0,\hat{\mathbb{P}}_0, \hat{\mathbb{Q}}_0)$. Observe that any sequence generated from $p(\cdot|x,p_0,\hat{\mathbb{P}}, \hat{\mathbb{Q}})$ is also admissible for $p(\cdot|x,p_0,\hat{\mathbb{P}}_0, \hat{\mathbb{Q}}_0)$ and vice versa. Therefore we generated 3000 sequences from $p(\cdot|x,p_0,\hat{\mathbb{P}}, \hat{\mathbb{Q}})$, another 3000 sequences from $p(\cdot|x,p_0,\hat{\mathbb{P}}_0, \hat{\mathbb{Q}}_0)$, and  put these two sets together to obtain a final set of 6000 initial sequences for case 1a and case 1b. The reason for using different parameter matrices is that sequences generated from one model tend to be alike, thus merging the two sets increases variety of initial sequences.
For case 2a and case 2b the set of initial sequences was obtained similarly: we generated 3000 sequences from  $p(\cdot|x,p_0,\hat{\mathbb{P}},\hat{\mathbb{Q}})$
and 3000 sequences from $p(\cdot|x,p_0,\hat{\mathbb{P}}_0,\hat{\mathbb{Q}}_0)$, and joined these sets into a set of 6000 initial sequences. Thus, we use one set of initial sequences in case 1a and case 1b, and another one in case 2a and case 2b.
\\\\
The segmentation results for case 1a  are presented in Table \ref{Ex1Case1}, the summary of the best paths is given in Table \ref{Ex1Case1paths}. The results for case 1b are presented
in Tables \ref{Ex1Case2} and  \ref{Ex1Case2paths}. The segmentation results with some path characteristics for case 2a are presented in Tables \ref{Ex1Case3} and \ref{Ex1Case3paths},
and the results for case 2b are presented in Tables \ref{Ex1Case4} and \ref{Ex1Case4paths}.

\begin{table}[htbp!]
{\scriptsize
\begin{tabular}{|c  c| l  | l  | l  | l  | l | l | }
\hline
   $N$ & $M$ &   sEM &  sMM &  VB &  BEM & Viterbi & Path$_0$ \\
  \hline  \hline
$20n$ &  $20n$  &   {\bf -973.72} (2) [2] & {\bf -973.72} (2) [2]  &{\bf -973.72} (1) [3] &{\bf -973.72} (1) [3] &{\bf -973.72} &  -996.13 \\
$2n$ & $2n$  &   {\bf -972.74} (1) [3]    & {\bf -972.74} (1) [3]  &-974.01 (2) [2]       &-974.01 (2) [2]       &-974.05       &  -998.01 \\
 $n$ & $n$  &   {\bf -971.94} (1) [3]     & {\bf -971.94} (1) [3]  &-974.06 (9) [2]       &-974.75 (10) [3]      &-974.83       &  -999.85 \\
$N_1+1$ & $M_1+1$ &   {\bf -971.29} (2) [3] &    -971.76 (3) [3]    &-973.37 (32)[2]      &-973.37 (45) [2]      &-975.96       &  -1001.91 \\
$n/2$ &  $n/2$  &   {\bf -971.26} (1) [4]   &  na                   & -973.75 (51) [2]    &     na                & -976.69   &   -1003.01  \\
$N_1/2$ &  $M_1/2$ &   {\bf -971.68} (1) [3] &   na                   &-974.22 (104) [2]   &     na                & -979.02   &  -1006.35   \\
$N_1/4$ &  $M_1/4$ &   {\bf -972.64} (1382) [4] &  na                   &-974.30 (220) [2]   &     na                & -984.23   &   -1013.07  \\
\hline
$20n$  &  $2n$  &   {\bf -975.64} (11) [3]  &{\bf -975.64} (13) [3] &{\bf -975.64} (2) [3] &{\bf -975.64} (2) [3] & -975.65 &     -998.02  \\
$20n$  &  $n$  &     {-977.33} (33) [3]      & {\bf -977.33} (37) [2]     &-977.47 (4) [3]    & -977.47 (10) [3]        & -977.51 &     -999.80   \\
$20n$ &  $M_1+1$ &   {\bf -979.01} (81) [2]      & {\bf -979.01} (67) [2] &-979.36 (22) [2]     & -979.40 (35) [2]       & -979.51  &    -1001.71  \\
$20n$ &  $n/2$  &  {\bf -979.93} (96) [3]   &   na                  &   -980.26 (40) [2] &  na                    & -980.56 &    -1002.72  \\
$20n$ &  $M_1/2$ &  {\bf -982.66} (145) [3] &   na                  &  -983.08 (96) [3]  &  na                    & -983.61 &    -1005.70  \\
$20n$ &  $M_1/4$ &  {\bf -987.71} (635) [4] &   na                  &  -987.86 (195) [2] &  na                    & -989.30 &    -1011.55 \\
\hline
$2n$ &   $20n$ &   {\bf -971.60} (1) [3]    & {\bf -971.60} (1) [3] & -972.13 (1) [3]       &-972.10 (2) [2]      & -972.13 &   -996.12    \\
$n$ &   $20n$ &   {\bf -969.37} (1) [3]     & {\bf -969.37} (1) [3] & -971.00 (3) [2]       &-971.00 (2) [2]      & -971.04 &   -996.19    \\
$N_1+1$ & $20n$ &   {\bf -967.51} (1) [3]   & {\bf -967.51} (2) [3] & -970.12 (3) [2]       &-969.52 (3) [2]      & -970.18 &   -996.33    \\
$n/2$ &  $20n$ &    {\bf -966.76} (1) [3]   & na                    &-969.08 (4) [2]       &  na                    & -969.85 &   -996.42   \\
$N_1/2$ & $20n$ &    {\bf -964.92} (1) [3]   & na                    &-968.09 (4) [2]      &  na                    & -969.14 &   -996.78   \\
$N_1/4$ & $20n$ &    {\bf -962.58} (1) [4]   & na                    &-967.14 (10) [2]      &  na                    & -968.65 &   -997.65   \\
\hline
\end{tabular}}
\caption{\label{Ex1Case1} (Case 1a) Parameter matrices $\hat{\mathbb{P}}$ and $\hat{\mathbb{Q}}$ have been estimated using the pair $(x,y)$ with $n=327$, 3000 initial path sequences
   have been generated from $p(\cdot|x,p_0,\hat{\mathbb{P}},\hat{\mathbb{Q}})$ and 3000 initial path sequences have been generated from $p(\cdot|x,p_0,\hat{\mathbb{P}}_0,\hat{\mathbb{Q}}_0)$ (see case 1b). The concentration parameters $N$ and $M$ have been tuned with respect to $N_1=198$ and $M_1=200$, $\alpha=N \hat{\mathbb{P}}$, $\beta=M \hat{\mathbb{Q}}$. The log-likelihood value $\ln p(\hat{v},x)$ of the estimated Bayesian Viterbi path for each algorithm and each set of hyperparameters is presented in the table, the number of distinct output sequences is given in round brackets, whereas the number of iterations for the best path is given in squared brackets. The log-likelihood of the Viterbi path calculated with $(p_0, \hat{\mathbb{P}},\hat{\mathbb{Q}})$ and the log-likelihood for the best initial path are also presented in the table.}
\end{table}
\begin{table}[htbp!]
{\footnotesize
\begin{tabular}{|c  c| c  | c  | c  | c  | c | c || c | c | c | c | c | c ||c|}
\hline
  $N$ & $M$ &   sEM &  sMM &  VB &  BEM & Viterbi & Path$_0$  & 1 & 2 & 3 & 4 & 5 & 6 & Blocks \\
\hline  \hline
$20n$ &  $20n$  &   0  &   0  &   0  &   0  &   0  &  64  &     13  &  25 &  263  &   6  &   8  &  12  & 17  \\
$2n$ & $2n$     &   0  &   0  &  32  &  32  &  34  &  79  &     13  &  17 &  297  &      &      &   & 7  \\
 $n$ & $n$      &   0  &   0  &  26  &  32  &  34  &  79  &     13  &  17 &  297  &      &      &   & 7  \\
$N_1+1$ & $M_1+1$ &  0  &  7  &  18  &  18  &  41  &  86  &     13  &  10 &  304  &      &      &   & 5  \\
$n/2$ &  $n/2$  &    0  &  na   &  18 &  na   &  41  &   86    &    13  &  10 &  304  &  &    &  & 5 \\
$N_1/2$ &  $M_1/2$ &  0 &  na  &   13  & na   &  41 &   86    &   13  &  10 &  304   &  &    &  & 5   \\
$N_1/4$ &  $M_1/4$ &  0 &  na &   10  &  na  &  51  &  96    &   13  &    &  314   &  &    &    & 3 \\
\hline
$20n$  &  $2n$  &   0  &   0  &   0  &   0  &   2  &  62     &   13  &  23 &  265  &   6   &  8  &  12 & 17  \\
$20n$  &  $n$  &    6  &  0   &  32  &  32  &  34  &  79    &   13   &  17 &  297  &   &   &      & 7    \\
$20n$ &  $M_1+1$ &  0  &   0  &  28  &  31  &  34  &  79    &   13   &  17  & 297  &   &   &      & 7    \\
$20n$ &  $n/2$  &  0  &  na   &  17  &  na  &  34  &  79      &  13  &  17  & 297  &       &      &   & 7    \\
$20n$ &  $M_1/2$ &   0  &  na   &  11  &  na  &  28  &  73    &  13  &  23  & 291  &       &      &   & 9   \\
$20n$ &  $M_1/4$ &   0  & na  &   11 & na   &   28 &   73   &   13 &   23 &  291 &       &      &     & 9 \\
\hline
$2n$ &   $20n$ &   0  &   0  &  34  &  32  &  34  &  79      &   13  &  17 &  297 &       &      &    & 7  \\
$n$ &   $20n$ &    0  &   0  &  32  &  32  &  34  &  79      &   13  &  17 &  297 &       &      &    & 7  \\
$N_1+1$ & $20n$ &    0  &   0  &  32  &  26  &  34 &   79    &  13  &  17 &  297  &       &      &    & 7  \\
$n/2$ &  $20n$ &   0  &  na   &  26  &  na  &  34  &  79       &  13  &   17 &  297  &     &      &   & 7  \\
$N_1/2$ & $20n$ &   0 &  na    &  26 &  na   &  34 &   79      &  13  &  17 &  297   &     &      &    & 7 \\
$N_1/4$ & $20n$ &   0  & na   &   28 &  na   &   41  &  86      &  13  &  10 &  304   &     &      &   & 5  \\
\hline
\end{tabular}}
\caption{\label{Ex1Case1paths} (Case 1a) In columns sEM,\ldots,Path$_0$, pointwise differences between the estimated Viterbi paths and the best path (in this case always sEM except in one case) are given for each method and each set of hyperparameters. The next six columns present state frequencies of the best path, the paths with the same state frequencies are equal. The last column gives the number of blocks in the best state sequence.}
\end{table}
\begin{table}[htbp!]
{\scriptsize
\begin{tabular}{|c  c| l  | l  | l  | l  | l | l |}
\hline
   $N$ & $M$ &   sEM &  sMM &  VB &  BEM & Viterbi & Path$_0$ \\
\hline  \hline
$20n$ &  $20n$  &   {\bf -1080.21} (3) [3] &{\bf -1080.21} (3) [3] &-1080.23 (1) [3] &-1080.23 (1) [3]     &-1080.23   &      -1148.91  \\
$2n$ & $2n$   &   {\bf -1075.66} (32) [5] &{\bf -1075.66} (31) [5] &-1075.74 (3) [2] &-1075.74 (3) [2]   &-1076.21   &      -1147.79  \\
$n$ & $n$  &   {\bf -1071.32} (385) [5] &{\bf -1071.32} (450) [5] &-1071.69 (7) [6] &-1071.69 (6) [6] &-1072.64   &      -1147.01  \\
 $n/2$ &   $n/2$   &   -1063.76 (2150) [5] & {\bf -1063.73} (2465) [6] & -1065.58 (25) [2] &-1065.58 (70) [2] &-1067.41  &      -1146.36  \\
$ n/4$ &   $n/4$   &  {\bf -1051.64} (3609) [9] & {\bf -1051.64} (4271) [5] &-1055.84 (149) [3] &-1055.59 (169) [3] & -1061.36  &   -1122.01 \\
$N_1+1$  &   $M_1+1$  &  {\bf -990.67} (3009) [3] &{\bf -990.67} (5903) [3] &{\bf -990.67} (2416) [6] &-1010.68 (5130) [2] &-1055.73  & -1047.61 \\
\hline
 $20n$ &   $2n$   &  {\bf -1077.77} (15) [4] &{\bf -1077.77} (18) [4] &-1077.78 (2) [4] &-1077.78 (2) [4] &-1078.12     &      -1148.06  \\
 $20n$  &    $n$   &  {\bf -1075.58} (219) [4] &{\bf -1075.58} (257) [6] &-1075.69 (5) [4] &-1075.66 (7) [2] &-1076.33   &      -1147.49  \\
  $20n$  &   $n/2$  &  {\bf -1071.57} (2261) [3] &{\bf -1071.57} (2723) [3] &-1072.59 (25) [3] &-1072.59 (46) [3] &-1073.87   &  -1147.07  \\
  $20n$  &   $n/4$  &  -1063.13 (4379) [11] &{\bf -1062.95} (4900) [7] &-1066.79 (137) [3] &-1067.28 (183) [4] &-1071.49      &  -1147.78  \\
 $20n$   &   $M_1+1$ &  -1037.26 (5999) [10] &{\bf -1036.56} (5997) [9] &-1045.62 (3822) [11] &-1038.48 (5395) [21] &-1073.91  &  -1159.48  \\
\hline
 $2n$ &    $20n$   &   {\bf -1078.18} (13) [3] &{\bf -1078.18} (13) [3] &-1078.20 (1) [4] &-1078.20 (1) [4] &-1078.32  &  -1148.64  \\
  $n$ &    $20n$   &   {\bf -1076.11} (38) [4] &{\bf -1076.11} (39) [4] &-1076.22 (7) [2] &-1076.28 (5) [3] &-1076.54  &  -1148.44 \\
 $n/2$ &    $20n$  &   {\bf -1072.22} (129) [5] &{\bf -1072.22} (136) [5] &-1072.76 (2) [6] &-1072.76 (2) [6] &-1073.77 &  -1148.21 \\
 $n/4$ &   $20n$   &   {\bf -1062.61} (57) [5] &-1062.78 (62) [5] &-1067.48 (15) [2] &-1067.60 (17) [2] &-1070.11      &    -1133.32 \\
 $N_1+1$ &   $20n$  &   {\bf -992.83} (300) [2] &{\bf -992.83} (295) [2] &{\bf -992.83} (1031) [2] &{\bf -992.83} (1227) [2] &-1062.05  & -1057.64 \\
\hline
\end{tabular}}
\caption{\label{Ex1Case2} (Case 1b) The probability mass between the non-zero elements in each row in both parameter matrices from case 1a is divided evenly (the case of uninformative priors), $\alpha=N\hat{\mathbb{P}}_0$, $\beta=M\hat{\mathbb{Q}}_0$, $N_1=4$, $M_1=20$. The same 6000 initial paths have been used as in case 1a. The log-likelihood value
$\ln p(\hat{v},x)$ of the estimated Bayesian Viterbi path for each algorithm and each set of hyperparameters is presented in the table, the number of distinct outputs is given in round brackets, whereas the number of iterations for the best path is given in squared brackets. The log-likelihood of the Viterbi path calculated with $(p_0,\hat{\mathbb{P}}_0,\hat{\mathbb{Q}}_0)$ and the log-likelihood for the best initial path is also presented in the table.}
\end{table}
\begin{table}[htbp!]
{\footnotesize
\begin{tabular}{|c  c| c  | c  | c  | c  | c | c || c | c | c | c | c | c || c|}
\hline
  $N$ & $M$ &   sEM &  sMM &  VB &  BEM & Viterbi & Path$_0$  & 1 & 2 & 3 & 4 & 5 & 6 & Blocks\\
 \hline  \hline
$20n$ &  $20n$  &      0  &   0   &  7  &   7  &  12  & 143   &   73  &  16 &  69 &  31 & 76 & 62  &  166 \\
$2n$ &  $2n$  &      0  &   0   &  2  &   2  &  19  & 143   &   70  &  14 &  67 &  32 & 76 & 68  &  165 \\
$n$ &  $n$  &     0   &  0  &  19  &  19  &  36  & 145   &   68  &  4  &  67  & 37 & 79 & 72  &  171  \\
$n/2$ &   $n/2$   &     1  &   0  &  54  &  54 &   62  & 145   &   68 &   3  &  70  & 43 &  86 & 57  &  194  \\
 $n/4$ &   $n/4$   &     0   &  0  &  52  &  65  &  69 &  241  &    66  &  3  &  71 &  43 & 88 & 56  &  197  \\
$N_1+1$ &   $M_1+1$  &      0  &   0  &   0  &  34  & 244  &  59  &    13 &      &  314 &   &    &     &   3   \\
\hline
 $20n$ &   $2n$   &     0   &  0  &   2   &  2  &  14  & 141  &    73  &  14  & 69  & 32 & 76 & 63  &  167  \\
 $20n$  &    $n$   &    0   &  0  &  13  &   8  &  25 &  144  &    71  &  10 &  69  & 32 & 77 & 68  &  168  \\
 $20n$  &    $n/2$  &   0   &  0  &  47  &  47  &  55 &  164  &    81  &  19 &  75  & 31 & 68 & 53  &  171  \\
 $20n$  &    $n/4$  &    59  &  0  &  70  &  65  & 100 &  181  &    63  &  27 &  69 &  31 & 68 & 69  &  165  \\
 $20n$ &   $M_1+1$ &  131  &   0  &  95  &  22  & 168 &  187  &    44  &  45 &  87 &  32 & 53 & 66  &  182  \\
\hline
 $2n$  &    $20n$   &     0  &   0  &   2  &   2  &  14  & 141  &   73  &  14 &  69  & 32  &  76 &  63 &  167  \\
 $n$  &    $20n$  &    0   &  0  &  30  &  20  &  29 &  146   &   71  &  11 &  69  & 35  & 83 & 58  & 172   \\
 $n/2$  &    $20n$ &     0   &  0  &  58  &  58 &   87 &  158  &    48 &   20 &  52  & 34 &  78 &  95 &  147  \\
 $n/4$  &    $20n$   &     0  &  20  & 148  & 131 &  149 &  258  &    45  & 133 &  40 &  18 &  54 & 37 &  94  \\
 $N_1+1$ & $20n$  &    0  &   0   &  0   &  0  & 244  &  87  &    13   &    &  314 &    &     &    &   3   \\
\hline
\end{tabular}}
\caption{\label{Ex1Case2paths} (Case 1b) In columns sEM,\ldots,Path$_0$, pointwise differences between the estimated Viterbi paths and the best path are given for each method and for each set of hyperparameters. The next six columns present state frequencies of the best path for each set of hyperparameters. The last column gives the number of blocks in the best state sequence.}
\end{table}

\begin{table}[htbp!]
{\scriptsize
\begin{tabular}{|c  c| l  | l  | l  | l  | l | l | }
\hline
   $N$ & $M$ &   sEM &  sMM &  VB &  BEM & Viterbi & Path$_0$ \\
\hline  \hline
$20n$ &  $20n$  &  {\bf -1013.86} (1) [2]     & {\bf -1013.86} (1) [2]    &{\bf -1013.86} (1) [2]       &{\bf -1013.86} (1) [2] & {\bf -1013.86}   &   -1063.26 \\
$N_1+1$ &  $n$  &  {\bf -1002.72} (15) [3]    & {\bf -1002.72} (15) [4]   &   -1003.94  (2) [2]         &   -1003.94  (3) [2]   & -1004.07      &       -1062.53  \\
$N_1+1$ & $M_1+1$ &  {\bf -1002.68} (43) [3]  & {\bf -1002.68}  (51)[4]   &  -1004.13  (3) [2]          & -1004.13  (6) [2]   & -1004.27    &         -1063.15 \\
$n/2$ &  $n/2$  &  {\bf -988.93}  (171) [3]   &         na                &  {\bf -988.93}  (9) [2]     &         na             &  -992.17  &        -1062.22 \\
$n/4$ &  $n/4$  &  {\bf -983.54} (473) [3]    &         na                &  {\bf -983.54} (139) [4]    &         na         &     -987.85     &       -1065.25 \\
$n/8$ &  $n/8$  &  {\bf -981.20} (1389) [4]   &         na                &  {\bf -981.20} (1863) [5]   &         na         &     -986.40     &       -1072.12 \\
\hline
$20n$ &  $n$   & {\bf -1011.39} (17) [3]    & {\bf -1011.39} (23) [3]  & -1011.72 (2) [2]          & -1011.72  (2) [2]      &   -1011.77   &         -1063.75 \\
$20n$ & $M_1+1$  & {\bf -1011.34} (40) [3]   & {\bf -1011.34} (81) [3]  &     -1011.91 (2) [2]      &  -1011.91  (4) [2]     &  -1011.97    &     -1064.38 \\
$20n$ & $n/2$  & {\bf -1011.70} (193) [5]   &            na        &   {\bf -1011.70} (9) [2]      &   na                   &    -1012.54   &    -1065.42 \\
$20n$ &  $n/4$  & {\bf -1012.65} (1827) [3]   &            na        &   {\bf -1012.65} (218) [3]   &   na                   &    -1014.96  &    -1069.23 \\
$20n$ &  $n/8$  &  {\bf  -1013.43} (4464) [7] &      na       &     -1014.07 (2838) [5]             &   na                   &     -1019.19    &         -1076.32  \\
\hline
$N_1+1$ & $ 20n$ & {\bf -1006.16}  (1) [2]   & {\bf -1006.16} (1) [2]  & {\bf -1006.16} (2) [2]  & {\bf -1006.16} (2) [2]  & {\bf -1006.16}   &  -1062.03 \\
 $n$ &  $20n$ &  {\bf -999.79}  (1) [3]   &       na                  &      -1000.25  (2) [2]  &      na          &  -1000.25    &        -1061.08 \\
 $n/2$ & $20n$ &  {\bf -992.28}  (1) [3]  &       na                    &      -993.50  (3) [2]  &        na           &  -993.50    &        -1060.06 \\
 $n/4$ & $20n$ &  {\bf -984.75}  (3) [3] &        na                 &         -986.75  (6) [2]   &       na           &  -986.75   &     -1059.27 \\
 $n/8$ & $20n$ &  {\bf -978.30}  (4) [3]    &         na              &   {\bf -978.30} (16) [2]   &        na          &   -981.07   &     -1059.06   \\
\hline
\end{tabular}}
\caption{\label{Ex1Case3} (Case 2a) The parameters $\hat{\mathbb{P}}$ and $\hat{\mathbb{Q}}$ have been estimated using the counts from the training data (1000 sequence pairs of arbitrary length). Now $N_1=682$, $M_1=231$.
Let $\hat{\mathbb{P}}_0$ and $\hat{\mathbb{Q}}_0$ be matrices obtained from $\hat{\mathbb{P}}$ and $\hat{\mathbb{Q}}$ by dividing the probability mass in each row evenly between the nonzero elements.
Again 3000+3000=6000 initial paths have been generated from the posterior distributions $p(\cdot|x,p_0,\hat{\mathbb{P}},\hat{\mathbb{Q}})$ and $p(\cdot|x,p_0,\hat{\mathbb{P}}_0,\hat{\mathbb{Q}}_0)$. The log-likelihood value $\ln p(\hat{v},x)$ of the estimated Bayesian Viterbi path for each algorithm and each set of hyperparameters is presented in the table, the number of distinct output sequences is given in round brackets, whereas the number of iterations for the best state path is given in squared brackets. The log-likelihood of the Viterbi path calculated with $(p_0,\hat{\mathbb{P}},\hat{\mathbb{Q}})$ and the log-likelihood of the best initial path is also presented in the table.}
\end{table}
\begin{table}[htbp!]
{\footnotesize
\begin{tabular}{|c  c| c  | c  | c  | c  | c | c || c | c | c | c | c | c || c |}
\hline
  $N$ & $M$ &   sEM &  sMM &  VB &  BEM & Viterbi & Path$_0$  & 1 & 2 & 3 & 4 & 5 & 6 & Blocks \\
 \hline  \hline
$20n$ &  $20n$  &    0  &   0   &  0  &   0  &   0  &  96     &        &  13 &  314  &    &    &   & 3 \\
$N_1+1$ &  $n$  &     0  &   0  &  11  &  11  &  13 &  109    &        &   &  327  &    &    &  & \\
$N_1+1$ & $M_1+1$ &    0  &   0  &  11  &  11  &  13  & 109   &          &    & 327    &    &    & & \\
$n/2$ &  $n/2$  &    0  &  na  &   0  &  na  &  13  & 109      &          &    & 327   &    &    & & \\
$n/4$ &  $n/4$  &    0  &  na  &   0  &  na  &  13 &  109   &          &     & 327   &      &     &    &   \\
$n/8$ &  $n/8$  &   0  &  na  &   0  &  na  &  13 &  109      &         &     & 327  &     &     &     &   \\
\hline
$20n$ &  $n$   & 0  &   0  &  11  &  11  &  13  & 109        &         &    & 327  &       &     &   & \\
$20n$ & $M_1+1$ & 0  &   0  &  11  &  11  &  13  & 109        &         &    &  327 &       &     &   & \\
$20n$ & $n/2$  &  0 &   na  &   0  &  na  &  13 &  109       &         &    &  327 &       &     &   & \\
$20n$ &  $n/4$  &  0  &  na  &   0  &  na  &  88 &  152       &         &  75 &  252 &     &      &   & 7 \\
$20n$ &  $n/8$  &   0  &  na  &  52  &  na &  126 &  172      &         & 113 &  214 &     &      &   &  11\\
\hline
$N_1+1$ & $ 20n$ &  0   &  0  &   0  &   0  &   0  &  96      &         &  13 &  314 &      &     &  & 3  \\
 $n$ &  $20n$ &   0  &  na  &  13  &  na  &  13 &  109       &        &     &   327 &      &     &  &  \\
 $n/2$ & $20n$ &   0  &  na  &  13  &  na  &  13 &  109       &        &     & 327   &      &     &  &  \\
 $n/4$ & $20n$ &   0  &  na  &  13  &  na  &  13 &  109       &        &     & 327   &      &     &  &  \\
 $n/8$ & $20n$ &    0  &  na  &   0 &   na  &  13 &  109      &        &     & 327   &      &     &  &   \\
\hline
\end{tabular}}
\caption{\label{Ex1Case3paths} (Case 2a) In columns sEM,\ldots,Path$_0$, pointwise differences between the estimated Viterbi paths and the best path are given for each method and for each set of hyperparameters. The next six columns present state frequencies of the best path for each set of hyperparameters. The last column gives the number of blocks in the best state sequence.}
\end{table}
\begin{table}[htbp!]
{\scriptsize
\begin{tabular}{|c  c| l  | l  | l  | l  | l | l | }
\hline
   $N$ & $M$ &   sEM &  sMM &  VB &  BEM & Viterbi & Path$_0$ \\
  \hline  \hline
 $20n$  &  $20n$  &   {\bf -1198.07} (1) [2] & {\bf -1198.07} (1) [2] &{\bf -1198.07} (1) [2] &{\bf -1198.07} (1) [2] &{\bf -1198.07} &    -1341.15 \\    
 $2n$  &  $2n$  &     {\bf -1145.47} (1) [2] & {\bf -1145.47} (1) [2] &{\bf -1145.47} (2) [2] &{\bf -1145.47} (2) [2] &{\bf -1145.47} &    -1320.91 \\    
 $n$  &  $n$  &      {\bf -1112.91} (3) [3] & {\bf -1112.91} (5) [3] &{\bf -1112.91} (2) [3] &{\bf -1112.91} (2) [3] &{\bf -1112.91} &    -1298.09 \\     
 $n/2$ &  $n/2$  &   {\bf -1077.48} (20) [4] &{\bf -1077.48} (24) [4] &{\bf -1077.48} (2) [3] &{\bf -1077.48} (2) [3] &{\bf -1077.48} &   -1254.52 \\     
 $n/4$ &  $n/4$ &    {\bf -1045.69} (43) [3] &{\bf -1045.69} (100) [3] &{\bf -1045.69} (29) [3] &{\bf -1045.69} (17) [5] &{\bf -1045.69} & -1209.95    \\  
 $n/8$ &  $n/8$  &   {\bf -1021.96} (205) [6] &{\bf -1021.96} (1470) [7] &{\bf -1021.96} (163) [7] &{\bf -1021.96} (935) [9] &{\bf -1021.96} & -1174.40   \\ 
$N_1+1$ & $M_1+1$ &   {\bf -986.61} (2057) [4] &-991.61 (5999) [2]   &  {\bf -986.61} (2716) [3]       &-1008.20 (6000) [3]      &{\bf -986.61}  &   -1117.14 \\ 
\hline
 $20n$  & $2n$   &  {\bf -1192.40} (1) [2]   &{\bf -1192.40} (1) [2] &{\bf -1192.40} (2) [2]    &{\bf -1192.40} (2) [2] &{\bf -1192.40}  &  -1339.85 \\  
 $20n$  & $n$   &  {\bf -1189.42} (2) [2]   &{\bf -1189.42} (2) [2] &{\bf -1189.42} (2) [2]    &{\bf -1189.42} (2) [2] &{\bf -1189.42}  &  -1338.47 \\   
 $20n$  & $n/2$  &  {\bf -1187.06} (2) [2]   &{\bf -1187.06} (2) [2] &{\bf -1187.06} (2) [2]    &{\bf -1187.06} (2) [2] &-1187.06        &  -1334.28 \\  
 $20n$  & $n/4$  &  -1184.84 (552) [6] &{\bf -1183.16} (3486) [13] &-1186.43 (7) [3] &-1186.43 (4) [3] &-1186.45    &   -1330.69  \\   
 $20n$  & $n/8$  &  -1146.64 (5997) [17] &{\bf -1144.39} (6000) [11] &-1188.14 (57) [6] &-1188.14 (1081) [7] &-1188.17      &   -1330.27 \\ 
 $20n$  & $M_1+1$ &  -1106.82 (6000) [11] &-1114.97 (6000) [10] &-1131.12 (5972) [27] &{\bf -1098.55} (6000) [22] &-1191.98    & -1335.69  \\ 
\hline
 $2n$  & $20n$ &  {\bf -1151.14} (1) [2]    &{\bf -1151.14} (1) [2] &{\bf -1151.14} (2) [2]    &{\bf -1151.14} (2) [2] &{\bf -1151.14}  & -1323.43 \\ 
 $n$  & $20n$ &  {\bf -1121.56} (2) [2]    &{\bf -1121.56} (2) [2] &{\bf -1121.56} (2) [2]    &{\bf -1121.56} (2) [2] &{\bf -1121.56}  &  -1305.07  \\ 
 $n/2$ & $20n$ &  {\bf -1088.49} (3) [2]    &{\bf -1088.49} (3) [2] &{\bf -1088.49} (2) [3]    &{\bf -1088.49} (2) [3] &{\bf -1088.49}  &  -1265.43  \\ 
 $n/4$ & $20n$ &  {\bf -1057.30} (5) [2]    &{\bf -1057.30} (5) [2] &{\bf -1057.30} (3) [3]    &{\bf -1057.30} (3) [3] &{\bf -1057.30}  &   -1222.16  \\ 
 $n/8$ & $20n$ &  {\bf -1031.86} (8) [2]    &{\bf -1031.86} (9) [2] &{\bf -1031.86} (5) [2]    &{\bf -1031.86} (5) [2] &{\bf -1031.86}  &   -1184.44 \\ 
 $N_1+1$ & $20n$ &  {\bf -992.70} (18) [2]     &{\bf -992.70} (29) [2]  &{\bf -992.70} (58)  [2]  &{\bf -992.70} (79) [2]  &{\bf -992.70} &   -1120.28 \\ 
\hline
\end{tabular}}
\caption{\label{Ex1Case4} (Case 2b) The probability mass of nonzero elements of the parameter matrices from case 2a is evenly distributed giving us $\hat{\mathbb{P}}_0$ and $\hat{\mathbb{Q}}_0$ (the case of uninformative priors), $\alpha=N\hat{\mathbb{P}}_0$, $\beta=M\hat{\mathbb{Q}}_0$, $N_1=4$, $M_1=20$. The same 6000 initial state paths have been used for segmentation as in case 2a. The log-likelihood value $\ln p(\hat{v},x)$ of the estimated Bayesian Viterbi path for each algorithm and each set of hyperparameters is presented in the table, the number of distinct output sequences is given in round brackets, whereas the number of iterations for the best state path estimate is given in squared brackets. The log-likelihood of the Viterbi path calculated with $(p_0, \hat{\mathbb{P}}_0,\hat{\mathbb{Q}}_0)$, and the log-likelihood with the best initial path is also presented in the table.}
\end{table}
%
%
%
%
%
\begin{table}[htbp!]
{\footnotesize
\begin{tabular}{|c  c| c  | c  | c  | c  | c | c || c | c | c | c | c | c ||c|}
\hline
  $N$ & $M$ &   sEM &  sMM &  VB &  BEM & Viterbi & Path$_0$  & 1 & 2 & 3 & 4 & 5 & 6 & Blocks \\
 \hline  \hline
 $20n$  &  $20n$  &   0  &    0   &   0  &    0  &    0  &  264    &      &      &    1  &    1  &  325  &   & 3 \\
 $2n$  &  $2n$  &   0  &    0  &    0   &   0  &    0  &   248      &      &      &    1  &    1  &  325  &  & 3  \\
 $n$  &  $n$  &   0  &    0  &    0  &    0   &   0  &  248        &       &     &    1  &    1  &  325  &   & 3 \\
 $n/2$ &  $n/2$  &  0  &    0  &    0   &   0   &   0  &  311      &        &      &   1  &    1 &   325  &   & 3\\
 $n/4$ &  $n/4$ &    0  &    0  &    0  &    0  &    0 &   311     &       &     &    1  &    1  &  325  &    & 3 \\
 $n/8$ &  $n/8$  &   0  &    0  &    0   &   0   &   0  &  311     &      &     &    1   &   1  &  325  &   & 3  \\
$N_1+1$ & $M_1+1$ &   0  &  326  &  0  &  325  &    0  &  311    &      &       &   1  &    1  &  325  &    & 3 \\
\hline
 $20n$  & $2n$   &   0   &   0  &    0   &   0   &   0 &   264     &       &     &    1  &    1  &  325  &   & 3 \\
 $20n$  & $n$   &   0  &    0   &   0   &   0   &   0  &  242      &      &       &   1  &    1  &  325  &   & 3 \\
 $20n$  & $n/2$  &    0  &    0  &    0  &    0   &   1  &  242    &      &      &    1  &    1  &  324  &    1 & 4 \\
 $20n$  & $n/4$  &  148  &    0 &   103  &  103  &  104  &  253    &   3  &   13  &   16  &   35 &   222  &   38 & 102\\
 $20n$  & $n/8$  &  270  &  0   &   253  &  253  &  254 &   269    &   43  &  51  &   51  &   56  &  72 &    54 & 220 \\
 $20n$  & $M_1+1$ & 267 &   244 &   297  &    0  &  283 &   266    &   58  &  61  &   71 &   50  &   43  &   44  & 225  \\
\hline
 $2n$  & $20n$ &    0  &    0   &   0   &   0  &    0  &  236     &       &      &    1  &    1 &   325  &  & 3  \\
 $n$  & $20n$ &   0  &    0  &    0   &   0  &    0  &  260        &       &     &    1  &    1 &   325  &  & 3   \\
 $n/2$ & $20n$ &   0  &    0 &     0  &    0  &    0 &   311       &      &      &    1  &    1  &  325  &  &  3  \\
 $n/4$ & $20n$ &   0  &    0   &   0  &    0  &   0  & 311       &       &     &   1  &   1  & 325  &   & 3 \\
 $n/8$ & $20n$ &   0  &   0  &   0   &  0   &  0  & 311       &       &     &  1   &  1  & 325  &   &  3   \\
 $N_1+1$ & $20n$ &   0  &   0  &   0  &   0  &   0  & 311     &       &    &   1  &   1  & 325  &   &  3  \\
 \hline
\end{tabular}}
\caption{\label{Ex1Case4paths} (Case 2b) In columns sEM,\ldots,Path$_0$, pointwise differences between the estimated Viterbi paths and the best path are given for each method and for each set of hyperparameters. The next six columns present state frequencies of the best path for each set of hyperparameters. The last column gives the number of blocks in the best state sequence.}
\end{table}
\subsubsection{Analysis of the results of the first experiments}
Before analyzing the results from case 1a, case 1b, case 2a and case 2b, recall that our goal is to solve the following optimization task with respect to different sets of hyperparameters $\alpha$ and $\beta$:
\begin{equation}\label{objective}
\arg\max_{s\in \{1,\ldots,K\}^n} \big[\ln p(x|s)+\ln p(s)\big],
\end{equation}
where $p(x|s)$ is defined as in (\ref{pxs}) (depending on  hyperparameters $\beta=M\mathbb{Q}$ and $x$) and  $p(s)$ is given in (\ref{ps}) (depending on
hyperparameters $\alpha=N\mathbb{P}$). Of course (\ref{objective}) also depends on fixed $p_0$. The function in (\ref{objective}) is a typical objective function in statistical learning, where $\ln p(x|s)$ is a data-dependent loss or risk function and $\ln p(s)$ is a regularization term. We already observed in Section \ref{sec:model} that in our case the regularization term aims to make the optimal paths more conservative (bigger blocks).
When $N$ (or $M$)  is very big (in our example $20n$), then the prior variances of the elements in the transition (or emission) matrix are so small that the matrix can be considered as fixed. Thus, a case with $N=20n$ corresponds to the model where the underlying process $Y$ is a Markov chain with the transition matrix $\mathbb{P}$, and a case with $M=20n$ corresponds to the case where given a state path $y$, the observations are independent with the emission matrix $\mathbb{Q}$. Fixing one of these matrices by taking the corresponding concentration parameter value high allows us to study the influence of the other parameter matrix, and this is why our numerical
 examples consist of three parts ($N=20n$ in the second part and $M=20n$ in the third part). In particular, the model with $M=N=20n$ is close  to an HMM with parameters $\mathbb{P}$ and $\mathbb{Q}$, and then it is clear that the path maximizing (\ref{objective}) is the Viterbi path obtained with  $\mathbb{P}$ and $\mathbb{Q}$, as all cases except case 1b also show.
\paragraph{Performance of different algorithms.} It is evident from Tables \ref{Ex1Case1}, \ref{Ex1Case2}, \ref{Ex1Case3} and \ref{Ex1Case4} that in most of the cases, sEM algorithm finds the best path and sMM (when applicable) performs very similarly. This is in full correspondence with the theory, because sEM algorithm is the only algorithm that increases the value of the objective function in (\ref{objective}). The similarity of sEM and sMM algorithms was shortly explained in Section \ref{sec:algorithms}. Observe that
 BEM and VB fail mostly to find the best path (except in case 2b). This is understandable, because the best path obtained with some parameter estimates is not  necessarily the best path when the parameters are integrated out. We also point out that even if the differences between the log-likelihoods of different paths are very small, it follows from Tables \ref{Ex1Case1paths}, \ref{Ex1Case2paths}, \ref{Ex1Case3paths} and \ref{Ex1Case4paths} that the paths can still be quite different. The log-likelihood of Path$_0$ shows that the best path cannot be obtained just by a lucky guess and all the algorithms actually improve the likelihood value.
\paragraph{Dependence on initial sequences.} The dependence on initial sequences grows when the concentration parameters decrease, i.e. when the prior variances increase and the influence of data decreases. The examples show that the dependence on  initial sequences is especially large in case 1b and case 2b, when the transition matrix is fixed ($N=20n$) and the emission prior variances are maximal ($M=M_1+1$). In this case basically every initial sequence produces a different output. Such a behaviour indicates that the likelihood function is in this case flat with many small local maxima, and we cannot be sure that the best path is the global maximum. The examples  confirm that when applying these iterative algorithms, the choice of initial sequences is crucial and cannot be overlooked.

\paragraph{Role of hyperparameters and structure of the estimated state paths.} Our examples demonstrate that the influence of hyperparameters on state path estimates is enormous: the hyperparameters influence the objective function (\ref{objective}) and the latter determines the structure of path estimates. Looking at cases 1a and 2a, we can see that the best state paths mostly consist of state 3, and in case 2a the path estimates often consist of only state 3. The paths dominated by one state in cases 1a and 2a have two explanations. First, the matrices $\hat{\mathbb{P}}$ and  $\hat{\mathbb{Q}}$ support in both cases the dominating state -- observe that state 3 is dominant even in the Viterbi path obtained with $\hat{\mathbb{P}}$ and  $\hat{\mathbb{Q}}$ (the first row in Table \ref{Ex1Case1paths} and \ref{Ex1Case3paths}). In case 2a, when the matrices $\hat{\mathbb{P}}$ and $\hat{\mathbb{Q}}$ have been estimated from training data and contain a smaller number of zeros compared to case 1a, the matrices seem to support a dominating state even more (compare the first rows in  Table \ref{Ex1Case1paths} and \ref{Ex1Case3paths}). The second reason is the influence of the regularization  term that prefers conservative paths. We can see that decreasing $N$ reduces the number of different states and blocks (Tables \ref{Ex1Case1paths}, \ref{Ex1Case2paths} and \ref{Ex1Case3paths}), just as the theory predicts.
In cases 1b and 2b  the matrices $\hat{\mathbb{P}}_0$ and  $\hat{\mathbb{Q}}_0$ are very different from $\hat{\mathbb{P}}$ and $\hat{\mathbb{Q}}$ in cases 1a and 2a, and obviously the structure of the best state paths is very  different as well. Observe also how different are the best paths in cases 1b and 2b. In case 1b the six states are more or less equally distributed, except the two cases with very small concentration parameter $N=N_1+1$ when the state 3 takes over (Table \ref{Ex1Case2paths}). The more equal state distributions are obtained because of uniform matrices $\hat{\mathbb{P}}_0$ and  $\hat{\mathbb{Q}}_0$, and the change for small $N$  clearly represents the strong influence of the regularization term. For case 2b the dominating state is 5. Recall that all the four cases study the same observation sequence $x$, hence the large variety of MAP path estimates shows that the influence of hyperparameters outperforms the influence of data.

\paragraph{States connected to certain observation letters.} Recall from Section \ref{sec:model} that putting the Dirichlet prior on emission probabilities has the effect of attaching every observation letter to one particular state. This tendency is more pronounced when the concentration parameter $M$ is small. Thus, for small $M$, the matrix $(m_{il}(x,y)): 6\times 20$ obtained with $x$ and MAP path estimate $y$ should be such that every column has a dominating element or quite many zeros. To see if we can observe this tendency in our example, we studied these matrices in case 1b when $N=20n$ (transition probabilities are fixed) and $M={2n}$,
$M=n$, $M={n/2}$, $M={n/4}$ and $M=M_1+1$. The total number of zeros in the matrices increased from 58 to 67. As a measure of sparseness, we calculated the entropy:
$H=-\sum_{i,j} (m_{ij}/n) \ln({m_{ij}/n})$. The entropy values for the given five values of $M$ are 3.85, 3.83, 3.80, 3.79 and 3.66, thus we can see that the entropy decreases as well.
%
\paragraph{Iterations.} The number of iterations needed for calculating the Bayesian Viterbi path estimates with our algorithms is typically below 10. The number of cases where more than 10 iterations were needed to find the optimal path is quite limited. This is good from practical point of view.
Observe also that even two iterations increase the path likelihood significantly, since the best initial path output (Path$_0$) is never comparable with the best output for any of the four studied methods.
\subsection{Frequentist, Bayesian and no training data approach}\label{sec:approaches}
We have a training set and an observation sequence $x$, which we assume to be an outcome of an HMM with an unknown parameter $\theta$.
Our goal is to find the Viterbi path $v(x):=\arg\max_{s}p(s|x,\theta)$. Since $\theta$ is unknown, there are in a large scale three
different approaches to solve the problem.
\paragraph{Frequentist approach.} In this case we assume that all training sequences $(x(k),y(k))$, $k=1,\ldots,m$, form an
 iid sample from the same HMM with the parameter $\theta^*$ (consisting of transition and emission matrices, because the initial distribution
is assumed to be known). Then also $x$ is an observation sequence from the same HMM and the solution to our problem is
straightforward: estimate the unknown parameters from the training data, let the
estimate be $\hat{\theta}$, and apply the Viterbi algorithm to find
$\arg\max_s p(s|x,\hat{\theta})$. It is reasonable to take
$\hat{\theta}=\big((\hat{p}_{ij}),(\hat{q}_{il})\big)$, where possible transitions and emissions
$\hat{p}_{ij}$ and  $\hat{q}_{il}$ are defined as previously by (\ref{point-est}), and for
impossible transitions and emissions the corresponding entries are zeros.
\paragraph{Bayesian approach.} Here we assume that the unknown
  parameter $\theta$ is random with distribution $\pi$. When we know
  $\pi$, then the best we can do is to find $\hat{y}$ that maximizes
  $$p(s|x)=\int p(s|x,\theta)p(d\theta|x)$$
  over all possible state paths $s$. In the Dircihlet case we have
  algorithms that perform the maximization. However, since we do not know the prior distribution, we
  use the training data to specify it. This approach
  assumes that every training pair $(x(k),y(k))$ is generated by a
  different parameter $\theta_k$, and these parameters can be considered as an iid
  sample from a common  prior $\pi(\theta)$. We consider Dirichlet priors, the hyperparameters are specified using the training set as explained in Section \ref{sec:emp}.
\paragraph{No training data case.} Suppose we do not believe that the training data are related to our
  $\theta$ or we believe that the sequence $x$ is long enough to estimate the unknown parameter $\theta$ solely based on $x$.
  In this case we can ignore the training data and apply the standard EM parameter estimation
 algorithm to $x$, obtain the parameter estimate $\hat{\theta}_{EM}$ and then apply the Viterbi algorithm to find
$\arg\max_s p(s|x,\hat{\theta}_{EM})$. When $x$ is long
enough, then due to posterior consistency and the consistency of
$\hat{\theta}_{EM}$, the Bayes approach and training data free
approach both yield the same result. For relatively short
sequences ignoring training data might be justified  by the
observation that the estimated variances in (\ref{variances}) are
relatively big.
\paragraph{Measure of goodness.} Suppose $\hat{v}$ is an estimate of the Viterbi path $v(x)$.
The correct measure of goodness of $\hat{v}$ is
$p(\hat{v}|x,\theta)$, but since  $\theta$ is unknown, one cannot
use this. Instead we have the true sequence $y$, that we can use for
testing purposes only. The most natural approach would be to use the
direct counts
\begin{equation}\label{counts}
\left({n_{ij}(y)\over n_i(y)}\right),\quad \left({m_{il}(x,y)\over
m_i(x,y)}\right)\end{equation} as the estimates of unknown parameters when validating the path estimates. But
since $x$ and $y$ might be very short, these
matrices might have too many zeros so that many admissible paths
would have probability zero when evaluated by these parameters.
Therefore we involve empirical priors (\ref{pr3}), (\ref{pr4}) and
find posteriors $p(\mathbb{P}|y)$ and $p(\mathbb{Q}|x,y)$. According
to (\ref{pos1}) and (\ref{pos2}) these posteriors factorize by rows
and the row posteriors (for possible transitions and emissions) are given by
\begin{align}\label{tr-pos1}
 (p_{i,1(i)},\ldots,p_{i,K_i(i)})\big|y&\sim \text{Dir}(N_i p^*_{i,1(i)}+n_{i,1(i)}(y),\ldots,N_i p^*_{i,K_i(i)}+n_{i,K_i(i)}(y)),\\
\label{tr-pos2}
(q_{i,1(i)},\ldots,q_{i,L_i(i)})\big|y,x& \sim \text {Dir}(M_iq^*_{i,1(i)}+m_{i,1(i)}(y,x),\ldots,M_iq^*_{i,L_i(i)}+m_{i,L_i(i)}(y,x)).
\end{align}
Denote the posterior mean matrices by $\bar{\mathbb{P}}=({\bar p}_{ij})$
and $\bar{\mathbb{Q}}=({\bar q}_{il})$, then for possible transitions and emissions,
\begin{equation}\label{bar}
{\bar p}_{ij}={N_ip^*_{ij}+n_{ij}(y)\over N_i+n_{i}(y)},\quad {\bar
q}_{il}={M_iq^*_{il}+m_{il}(x,y)\over M_i+m_{i}(x,y)}.\end{equation}
For impossible emissions and transitions the corresponding entries
are zeros. These matrices constitute the estimate $\bar{\theta}$
of the unknown parameter $\theta$. The reason for using posterior means
instead of posterior modes is that the latter might not be defined.
To be more representative, we consider a larger class of estimates
$\bar{\theta}^c$, where
\begin{equation}\label{barc}
{\bar p^c}_{ij}:={cN_ip^*_{ij}+n_{ij}(y)\over cN_i+n_{i}(y)},\quad
{\bar q^c}_{il}:={cM_iq^*_{il}+m_{il}(x,y)\over cM_i+m_{i}(x,y)},\end{equation} and $c\in (0,1]$. For small $c$,
the matrices $\big({\bar p^c}_{ij}\big)$ and $\big({\bar
q^c}_{il}\big)$ are close to counts (\ref{counts}), but all possible
transitions and emissions are positive. With parameters
$\bar{\theta}^c$, all admissible paths have positive posterior
probability, and therefore they can be evaluated. Thus, we have
specified our criterion of goodness: $p(\hat{v}|x,\bar{\theta}^c)$.
Since this probability depends much on the length of the sequences
and the length of the sequences in our data set varies a lot, we
use the geometric mean $p(\hat{v}|x,\bar{\theta}^c)^{1/n}$, where $n$ is the length of $x$ and $y$.\\\\
Recall once again that although it might be tempting to measure the goodness of obtained paths by calculating
the number of pointwise differences from the true state sequence $y$, this is not the right criterion, because the Viterbi path is not the best path for minimizing the expected number of errors.


\subsection{Example 2: comparison of the frequentist, Bayesian and no training data approach}
To compare and test the goodness of different approaches described, we proceed as follows. We consider a training set and test set both consisting
of $m=1000$ sequence pairs. The initial distribution $p_0$ is considered as fixed and calculated using the whole data set, it is given by
$p_0=(0.0016,0.0041,0.9929,0.0014,0,0)'$. We can see that most of the sequences start from state 3, very few sequences start from state 1, 2 or 4. Based on the training data, impossible transitions and emissions will be specified by simple counts. To compare the approaches, we calculate the following paths for every $x(k)$ in the test set:
\begin{description}
\item{\bf Frequentist approach:} the Viterbi estimate  $\hat{v}_{k}^1:=\arg\max_{s}p(s|x(k),\hat{\theta})$, where $\hat{\theta}$ is calculated from the training data using formulas (\ref{point-est}).
\item{\bf Bayesian approach:} the Viterbi paths are found using sEM and VB methods, denoted by
$\hat{v}^2_{k}$ and $\hat{v}^3_{k}$, respectively. For that, the empirical priors are calculated using the training set. For both algorithms 1000 initial sequences are used. These two algorithms were chosen because they are applicable for all the studied priors.
\item{\bf No trainig data approach:} the Viterbi path  $\hat{v}^{4}_k=\arg\max_{s}p(s|x(k),\hat{\theta}_{k})$, where
$\hat{\theta}_{k}$ is the parameter estimate obtained with the standard EM algorithm using solely $x(k)$. As initial parameter estimates in the EM algorithm
we use the direct counts (\ref{counts}) obtained with $x(k)$ and every initial state path. In addition, we consider the parameter estimates from the training set as
initial parameters. Thus, in total we run the EM algorithm 1001 times for every $x(k)$. The convergence criterion of the EM algorithm is determined through
the log-likelihood value.
\item{\bf Target Viterbi path:} all the path estimates defined in 1), 2) and 3) are supposed to estimate the true Viterbi path
$v_k=\arg\max_{s}p(s|x(k),\bar{\theta}^c_{k})$, where
$\bar{\theta}^c_{k}$ are the parameter estimates (\ref{barc})
obtained with $(x(k),y(k))$ and the training data. The  parameter $\bar{\theta}^c_{k}$ is considered as the true parameter for
$(x(k),y(k))$ and used in the criterion of goodness.
\end{description}
The goodness of the performance is measured via the following sums:
\begin{align*}
{\rm sum}(v)&:=\sum_k p(v_k|x(k),\bar{\theta}^c_{k})^{1/t_k},
\quad {\rm sum}(\hat{v}^j):=\sum_k p(\hat{v}_{k}^j|x(k),\bar{\theta}^c_{k})^{1/t_k},\quad j=1,\ldots,4,
\end{align*} where $t_k$ denotes the length of $x(k)$. Since for
every $k$ and for every state path $s$,
$p(v_k|x(k),\bar{\theta}^c_{k})\geq p(s|x(k),\bar{\theta}^c_{k})$,
the first sum --  ${\rm sum}(v)$ -- is clearly the biggest. This is
the benchmark for all the algorithms. To illustrate the differences compared to the largest sum more clearly, we present their relative differences (in percentage) as follows:
\begin{equation} \label{RelDiff1}
{{\rm sum}(\hat{v}^j)\over {\rm sum}({v})}\cdot 100, \quad j=1,\ldots,4.
\end{equation}
The other way for performing relative comparisons is to consider quantities
\begin{equation} \label{RelDiff2}
 {\rm mean}_{rel}(\hat{v}^j) =\frac{1}{m}\sum_{k=1}^m \Big[ {p(\hat{v}_{k}^j|x(k),\bar{\theta}^c_{k}) \over p(v_k|x(k),\bar{\theta}^c_{k})} \Big]^{1/t_k}, \quad j=1,\ldots,4.
\end{equation}
Recall that the constant $c$ is used to define the true
parameter $\bar{\theta}^c_{k}$, only. In the training algorithms the
original empirical hyperparameters are used. In what follows, we
 summarize the performance of the three approaches for three different subsamples of our data. \\\\
\textbf{Case 1: sequence pairs of similar length}. In the first example we consider 1000 sequence pairs of length between 180 and 220. The following behaviour of the obtained estimates can be
observed: sEM gives 889 constant sequences of state 3, VB gives 877 constant sequences of state 3, the sEM and VB estimates are equal in 939 cases out of 1000,
for the frequentist approach there are 61 constant estimates of state 3 and there are none for the EM case where the training data is not involved. These numbers illustrate very well the Bayesian effect -- the MAP paths tend to consist of very long blocks. Furthermore, in this example the path estimates are often constant sequences.
%
\begin{table}[htbp!]
\begin{center}
{\footnotesize
\begin{tabular}{| c | c | c | c | c |}
  \hline
$c$     &  Freq  & sEM  & VB  & EM   \\
  \hline
 $10^6$  &   100.0000 &  97.4706 &  97.4876 &  64.4912 \\
      1  &    91.5263 &  92.5350 &  92.6060 &  60.5583  \\
    0.8  &    89.8025 &  91.5393 &  91.6206 &  59.4776  \\
    0.6  &    87.3450 &  90.1529 &  90.2481 &  57.8715  \\
    0.4  &    83.5728 &  88.1220 &  88.2374 &  55.2762  \\
    0.3  &    80.7844 &  86.7031 &  86.8329 &  53.2629  \\
    0.2  &    76.8473 &  84.8193 &  84.9693 &  50.2833  \\
    0.1  &    70.4506 &  82.0179 &  82.1989 &  45.0820  \\
    0.005 &   51.4144 &  74.3159 &  74.5610 &  26.7992  \\
\hline
\end{tabular}
\begin{tabular}{| c | c | c | c | c |}
  \hline
$c$     &  Freq  & sEM  & VB  & EM   \\
  \hline
 $10^6$   &    1.0000 &   0.9747 &   0.9749 &  0.6449  \\
      1   &    0.9151 &   0.9251 &   0.9258 &  0.6057  \\
    0.8   &    0.8978 &   0.9151 &   0.9159 &  0.5949  \\
    0.6   &    0.8731 &   0.9012 &   0.9021 &   0.5788 \\
    0.4   &    0.8353 &   0.8808 &   0.8820 &   0.5528 \\
    0.3   &    0.8073 &   0.8666 &   0.8679 &   0.5327 \\
    0.2   &    0.7678 &   0.8477 &   0.8492 &   0.5030 \\
    0.1   &    0.7037 &   0.8196 &   0.8214 &   0.4511 \\
    0.005 &    0.5129 &   0.7424 &   0.7448 &   0.2688 \\
\hline
\end{tabular}}
\end{center}
\caption{\label{SimLength}(Case 1) Relative differences (on the left) calculated according to (\ref{RelDiff1}) and mean relative differences (on the right) calculated according to (\ref{RelDiff2}) for the frequentist method, sEM, VB and no training data case.}
\end{table}
%
%
\newline \newline
\textbf{Case 2: sequence pairs of arbitrary length}. In this example, the training set and test set consist both of 1000 randomly sampled sequence pairs. The obtained path estimates can be summarized as follows:
sEM gives 869 constant sequences of state 3, VB gives 861 constant sequences of state 3 as Viterbi estimates, in 961 cases out of 1000 the sEM and VB estimates are equal,
for the frequentist approach there are 189 constant estimates of state 3 and there are none for the EM case with no training data.
\begin{table}[htbp!]
\begin{center}
{\footnotesize
\begin{tabular}{| c | c | c | c | c |}
  \hline
 $c$    &  Freq  & sEM  & VB  & EM   \\
  \hline
 $10^6$  &  100.0000 &  98.0680 &  98.1113 &  65.5150  \\
      1  &   91.7996 &  92.1118 &  92.1989 &  61.2743  \\
    0.8  &   90.0150 &  90.7390 &  90.8326 &  60.0735  \\
    0.6  &   87.4541 &  88.7736 &  88.8762 &  58.2792  \\
    0.4  &   83.5158 &  85.7811 &  85.8971 &  55.3558  \\
    0.3  &   80.6174 &  83.6110 &  83.7367 &  53.0689  \\
    0.2  &   76.5702 &  80.6362 &  80.7756 &  49.7060  \\
    0.1  &   70.1373 &  76.0440 &  76.2063 &  43.9250  \\
    0.005 &  51.5464 &  63.2427 &  63.5506 &  24.4300  \\
\hline
\end{tabular}
\begin{tabular}{| c | c | c | c | c |}
  \hline
$c$     &  Freq  & sEM  & VB  & EM   \\
  \hline
 $10^6$  &    1.0000 &   0.9806  &   0.9811  &  0.6556 \\
      1  &    0.9179 &   0.9209  &  0.9217 &   0.6148  \\
    0.8  &    0.9000 &   0.9071  &  0.9081 &   0.6027  \\
    0.6  &    0.8743 &   0.8875  &  0.8885 &   0.5847  \\
    0.4  &    0.8349 &   0.8577  &  0.8588 &   0.5555  \\
    0.3  &    0.8059 &   0.8362  &  0.8374 &   0.5329  \\
    0.2  &    0.7656 &   0.8069  &  0.8083 &   0.4997  \\
    0.1  &    0.7017 &   0.7620  &  0.7637 &   0.4427  \\
    0.005  &  0.5175 &   0.6377  &  0.6403 &   0.2530  \\
\hline
\end{tabular}}
\end{center}
\caption{\label{ArbLength}(Case 2) Relative differences (on the left) calculated according to (\ref{RelDiff1}) and mean relative differences (on the right) calculated according to (\ref{RelDiff2}) for the frequentist method, sEM, VB and no training data case.}
\end{table}
%
%
\newline \newline
\textbf{Case 3: long sequence pairs}. As the last case, we compare the performance of the methods for the longest sequences. We consider 2000 longest sequence pairs splitted into a training and test set.
The characteristics of the obtained path estimates can be summarized as follows: sEM gives 817 constant sequences of state 3 as Viterbi estimates, VB gives 915 constant sequences of state 3, in 803 cases out of 1000 the sEM and VB estimates are equal,
for the frequentist approach there are 76 constant estimates of state 3 and for the no training data case there are none.
\begin{table}[htbp!]
\begin{center}
{\footnotesize
\begin{tabular}{| c | c | c | c | c |}
  \hline
 $c$    &  Freq  & sEM  & VB  & EM   \\
  \hline
 $10^6$  &   100.0000 &  98.3556 &  98.6987 &  61.4819 \\
   1   &   96.5499 &  95.8184 &  96.7807 &  59.3507  \\
  0.8  &   95.8726 &  95.3544 &  96.4418 &  58.7327  \\
  0.6  &   94.9013 &  94.7242 &  95.9852 &  57.8764  \\
  0.4  &   93.3812 &  93.8193 &  95.3423 &  56.4609  \\
  0.3  &   92.2193 &  93.1880 &  94.9041 &  55.3141  \\
  0.2  &   90.5125 &  92.2823 &  94.3393 &  53.5068  \\
  0.1  &   87.5456 &  91.0386 &  93.5532 &  49.8650  \\
 0.005 &   76.2530 &  87.3346 &  91.8399 &  34.3386  \\
\hline
\end{tabular}
\begin{tabular}{| c | c | c | c | c |}
  \hline
$c$     &  Freq  & sEM  & VB  & EM   \\
  \hline
 $10^6$   &  1.0000  &  0.9835 &   0.9869  &  0.6208 \\
    1   &   0.9654  &  0.9579  &  0.9677  &  0.6013  \\
   0.8  &   0.9586  &  0.9532  &  0.9642  &  0.5958  \\
   0.6  &   0.9488 &   0.9469  &  0.9596 &   0.5875  \\
   0.4  &   0.9335 &   0.9378  &  0.9531 &   0.5736  \\
   0.3  &   0.9218 &   0.9314  &  0.9487 &   0.5625  \\
   0.2  &   0.9046 &   0.9229  &  0.9430 &   0.5455  \\
   0.1  &   0.8747 &   0.9103  &  0.9351 &   0.5142  \\
  0.005 &   0.7643 &   0.8764  &  0.9179 &   0.3860  \\
\hline
\end{tabular}}
\end{center}
\caption{\label{LongLength}(Case 3) Relative differences (on the left) calculated according to (\ref{RelDiff1}) and mean relative differences (on the right) calculated according to (\ref{RelDiff2}) for the frequentist method, sEM, VB and no training data case.}
\end{table}
\paragraph{Discussion of the results.} The main difference between the frequentist, Bayesian and no training data case is how we use the available information from the training data. In the frequentist approach  the uncertainty of the point estimates is not counted for when performing segmentation for the test set. If the variances of the prior distributions are very small, then the frequentist and Bayesian approach give similar path estimates. Thus, the relative performance of the Bayesian approach in comparison to the frequentist approach is determined by which prior distribution we use. If the concentration parameters are large, then the priors are concentrated over the point estimates. In Table \ref{concentr}, the concentration parameters corresponding to empirical priors are presented for our three examples. We can observe that $N_6$ is large for all subsamples, meaning that the prior distribution of row 6 in the transition matrix is very much concentrated over the point estimates of the transition probabilities from state 6. We can also observe that for long sequences in case 3 the concentration parameters are about twice as large as for case 1 and case 2, meaning that the empirical variances of the parameters are smaller in the case of longest sequence pairs, which makes also sense.
Tables \ref{SimLength}, \ref{ArbLength} and \ref{LongLength} show that in all the three cases (with sequence pairs of different length) the Bayesian approach outperforms all other approaches. Furthermore, from the two Bayesian segmentation algorithms, VB performs slightly better. This slightly contradicts the results of our first experiments, where we tested performance of the segmentation algorithms, but on the other hand the difference between the performance measures of sEM and VB is small.

\paragraph{Large vs small $c$.} When $c$ is very large, then ${\bar p^c}_{ij}$ and  ${\bar q^c}_{il}$ will be very close to $p^{\ast}_{ij}$ and  $q^{\ast}_{il}$, thus the Viterbi estimate with $({\bar p^c}_{ij},{\bar q^c}_{il})$ is expected to be very close to the frequentist estimate. This can be seen in all our tables, where we can see that the frequentist path estimates coincide with the Viterbi estimates obtained with $({\bar p^c}_{ij},{\bar q^c}_{il})$ when $c=10^6$. Very small $c$-values are closest to the situation of validation with 'true' parameter values, because then ${\bar p^c}_{ij}$ and  ${\bar q^c}_{il}$ are close to the point estimates of the parameters obtained with counts from $\{x(k),y(k)\}$. Thus, in this sense the last row of each table with $c=0.005$ is most interesting. The value $c=1$ corresponds to the case when we validate the results under the posterior means: we believe in our prior distributions and use the information from $\{x(k),y(k)\}$ to update the parameter distributions.
We can observe for all the four studied methods and for all the three subsamples that when $c$ decreases and therefore, the influence of $\{x(k),y(k)\}$ in $\bar{\theta}^c_{k}$ increases, then the difference between $p(\hat{v}^j_k|x(k),\bar{\theta}^c_{k})$ and $p(v_k|x(k),\bar{\theta}^c_{k})$ increases on average. The degree of this increase depends on sequence length. For example, for case 1 with sequence length around 200, the starting position (with $c=1$) for Freq, sEM and VB is about 0.92, but the relative difference measures for $c=0.005$ have decreased to 0.51, 0.74 and 0.74, respectively. For case 2 with sequences of different lengths (503 sequences are of length between 30 and 141) this decrease of average relative difference measure is even larger: the starting position for Freq, sEM and VB is again 0.92, but for $c=0.005$ this number has decreased to 0.52, 0.64 and 0.64, respectively. In case 3 when the test sequences are longer than 360, the change in average relative difference measure is smaller, indicating that longer sequences contain more information about transition and emission parameters. In case 3, the starting values (when $c=1$) for Freq, sEM and VB are 0.97, 0.96 and 0.97, these values decrease to 0.76, 0.88 and 0.92, respectively, when $c=0.005$.
\begin{table}[htbp!]
\begin{center}
{\footnotesize
\begin{tabular}{| c | c  c  c  c c c || c  c  c  c c c|}
  \hline
      &  $N_1$  & $N_2$  & $N_3$  & $N_4$ & $N_5$   & $N_6$  & $M_1$  & $M_2$  & $M_3$  & $M_4$ & $M_5$   & $M_6$ \\
  \hline
  Case 1 &  53  &   71  &   49   &   178   &   17  &  956  &  12  &   34  &   62  &   17  &    24  &  16   \\
  \hline
  Case 2 &  57  &   77  &   49   &  149   &    17  &  1544  & 13  &   33  &   58 &    15  &   22   &  15   \\
  \hline
  Case 3 &  100 &  120  &   94   &  344 &   42  &   3676  &   24  &   63  &  117 &   35   &   44   &  32   \\
\hline
\end{tabular}}
\end{center}
\caption{\label{concentr}Concentration parameters $N_i$ and $M_i$, $i=1,\ldots,6$, corresponding to the empirical priors calculated from our training data sets with 1000
sequence pairs of length between 180 and 220 (case 1), of arbitrary length (case 2) and with longest sequences (case 3).}
\end{table}
%
\subsection*{Acknowledgments}
This work was supported by the Estonian institutional research funding IUT34-5 and by the Estonian Research Council grant PRG865.
%
\bibliographystyle{plain}
\bibliography{bayesbib}

\end{document}